\documentclass[journal]{IEEEtran}
\usepackage{graphics} 
\usepackage{epsfig} 
\usepackage{times} 
\usepackage{amsmath} 
\usepackage{amssymb}  
\usepackage[linesnumbered,ruled]{algorithm2e}
\usepackage{multirow}
\usepackage{bm,cases}
\usepackage{flushend}
\usepackage{multicol}
\usepackage{cite}
\usepackage{booktabs}
\usepackage{threeparttable}
\usepackage{array}
\usepackage{mathtools}
\usepackage{url}
\usepackage{CJK}
\usepackage{color}

\usepackage{mathrsfs}
\usepackage{epstopdf}

\usepackage{tikz}
\usepackage{etoolbox}
\newcommand*{\circled}[1]{\lower.2ex\hbox{\tikz\draw (0pt, 0pt)%
		circle (.35em) node {\makebox[1em][c]{\small #1}};}}
\robustify{\circled}

\begin{document}
\title{Online Initialization and Extrinsic Spatial-Temporal Calibration for Monocular Visual-Inertial Odometry}

%
%
%


\author{Weibo~Huang, Hong~Liu, and Weiwei~Wan
	\thanks{
		W. Huang and H. Liu are with Key Laboratory of Machine Perception, Peking University, Shenzhen Graduate School, Beijing, 100871, China (e-mail: weibohuang@pku.edu.cn; hongliu@pku.edu.cn).
	}%
	\thanks{W. Wan is with School of Engineering Science, Osaka Univeristy, Osaka, 5608531, Japan. (e-mail: wan@sys.es.osaka-u.ac.jp).}
}

\markboth{Journal of \LaTeX\ Class Files,~Vol.~14, No.~8, August~2015}%
{Shell \MakeLowercase{\textit{et al.}}: Bare Demo of IEEEtran.cls for IEEE Journals}

\maketitle

\begin{abstract}

This paper presents an \textit{online initialization} method for bootstrapping the optimization-based monocular visual-inertial odometry (VIO). The method can online calibrate the relative transformation (\textit{spatial}) and time offsets (\textit{temporal}) among camera and IMU, as well as estimate the initial values of metric scale, velocity, gravity, gyroscope bias, and accelerometer bias during the initialization stage.
To compensate for the impact of time offset, our method includes two short-term motion interpolation algorithms for the camera and IMU pose estimation.
Besides, it includes a three-step process to incrementally estimate the parameters from coarse to fine.
First, the extrinsic rotation, gyroscope bias, and time offset are estimated by minimizing the rotation difference between the camera and IMU.
Second, the metric scale, gravity, and extrinsic translation are approximately estimated by using the compensated camera poses and ignoring the accelerometer bias.
Third, these values are refined by taking into account the accelerometer bias and the gravitational magnitude.
For further optimizing the system states, a nonlinear optimization algorithm, which considers the time offset, is introduced for global and local optimization.
Experimental results on public datasets show that the initial values and the extrinsic parameters, as well as the sensor poses, can be accurately estimated by the proposed method.
\end{abstract}


%
\IEEEpeerreviewmaketitle

\section{Introduction}
The monocular visual-inertial odometry (VIO) technology, which aims to compute the incremental sensor motion and the scene structure by fusing measurements from a camera and an inertial measurement unit (IMU), has become an active research topic in robotics and computer vision communities.
Since cameras and IMUs are both cheap, ubiquitous, small in size, low in power consumption, and mutually complementary, these two sensor types are ideal choices for VIOs.
The image contains a rich representation of the environment, which can be utilized to build sparse/dense maps and to estimate the camera poses up-to-scale.
Given an initial pose and velocity, the short-term rigid body motion can be accurately estimated by integrating the angular velocity and local linear acceleration measured by IMUs.
These complementary features make the visual-inertial setup suitable for many applications like unmanned aerial robots \cite{shen2017jfr, bloesch2015robust},  autonomous or semi-autonomous driving\cite{jones2011visual, huai2018robocentric}, 3D reconstruction\cite{Dong_2017_CVPR, yang2017real}, and augmented reality (AR)\cite{oskiper2012ar,li2017monocular}, etc.

The performance of monocular VIOs heavily relies on the accuracy of the initial values (including metric scale, velocity, gravity, gyroscope bias, and accelerometer bias) and the relative spatial and temporal parameters between camera and IMU.
The spatial parameters are the bridge of state transformation between the camera reference frame and IMU reference frame, while the time offsets are used to align different sensor streams.
To process the sensor measurements in an estimator, each camera image and IMU measurement is attached with a timestamp, which is taken either from the sensor itself or from the operating system (OS) of the computer receiving the data.
Due to the unsynchronized clocks, transmission delays, sensor response, and OS overhead, there always exists a latency between the actual sampling instant and the attached timestamp.
Since the latency is different for each sensor, the measurement streams from the camera and IMU are usually misaligned.
If the spatial and temporal parameters are not considered or incorrectly calibrated, the performance of mapping and navigation would be severely impacted.

In early studies, offline methods\cite{rehder2017camera,rehder2016extending,furgale2013unified,furgale2012continuous,maye2013self} were commonly used to obtain precise extrinsic spatial and temporal parameters.
These solutions require a professional user to carefully move the sensor suite in front of a stationary visual calibration target, which is time consuming and usually inaccessible in some cases, e.g., rescue missions.
To overcome these shortcomings, several online methods were developed in more recent studies (see section \ref{sec:related work}).
However, to the best of our knowledge, there is not much work in contemporary publications studied the estimation of extrinsic spatial-temporal parameters along with all the initial values for bootstrapping the optimization-based VIOs.
To this end, we here present an online method for calibrating the extrinsic spatial and temporal parameters, as well as estimating the initial values of velocity, gravity, visual metric scale, and IMU biases.

Specifically, the first contribution of this work is the \textit{short-term sensor motion interpolation} algorithm. Our approach assumes that the sensor suite moves in constant angular and linear velocities between two keyframe instants. This assumption is reasonable since the time interval between two consecutive keyframe instants is usually tens to hundreds of milliseconds, which can be considered as a short term interval.
As a result, we design two motion interpolation algorithms, i.e., the \textit{camera motion interpolation} and the \textit{IMU motion interpolation}, to interpolate the camera pose and IMU pose at an arbitrary intermediate time.
By representing the interpolation as a function of the unknown time offset and the metric scale, we can establish the transformation relationship between camera and IMU at any timestamp.

The second main contribution of this work is the \textit{nonlinear optimization-based algorithm} for global/local optimizing the IMU states (including position, rotation, velocity, and biases), the reconstructed map points, and the extrinsic spatial-temporal parameters.
The \textit{IMU preintegration error} and \textit{feature reprojection error} are both minimized.
By applying the IMU motion interpolation, the feature reprojection error is formulated as a function of map point position, IMU pose, IMU velocity, and extrinsic spatial-temporal parameters.

The third main contribution of this work is the \textit{three-step process} for estimating the extrinsic spatial-temporal parameters and the initial values in a coarse-to-fine manner.
This three-step process is an extension of our previous work \cite{huang2018online}.
In particular, the temporal misalignment between different sensor streams is considered in this work.
To be specific, the spatial rotation, time offset, and gyroscope bias are estimated by minimizing the rotation difference between camera and IMU in the first process.
In the second process, the scale factor, gravity, and spatial translation are approximately estimated by using the interpolated camera poses and ignoring the accelerometer bias.
In the third process, the values estimated in the second process are further refined by taking the accelerometer bias and the gravitational magnitude into account.
The result of the three-step process is provided as the initial estimate for the global nonlinear optimization.


The remaining part of the paper is organized as follows: Section \ref{sec:related work} reviews the related works.
Section \ref{sec:Preliminary} discusses the IMU model and the IMU preintegration theory.
Section \ref{sec:short-term sensor motion interpolation} introduces the short-term sensor motion interpolation algorithm, including the time offset model, camera and IMU motion interpolation algorithms, and the sensor transformation relationship.
Section \ref{sec:Visual-Inertial_State_Estimation} presents the nonlinear optimization with time offset.
Section \ref{sec:IMU Initialization and Extrinsic Spatial-Temporal Calibration} introduces the details of the three-step process.
Experiments and analyses are performed in Section \ref{sec:experiments and discussions}.
Conclusions are drawn in Section \ref{sec:Conclusions}.

\section{Related Work}
\label{sec:related work}

In recent years there have been excellent results in monocular visual-inertial odometry techniques. They can be categorized into filter-based and optimization-based approaches based on the sensor fusion algorithm.
Filter-based approaches \cite{mourikis2007multi, li2012improving, tanskanen2015semi, bloesch2017iterated} generally employ Kalman filter or extended Kalman filter (EKF), and are suitable for computing resource-constrained platforms. In these approaches, the state propagation/prediction is made by integrating IMU measurements, and the update/correction is performed by using visual measurements.
On the contrary, typical optimization-based approaches \cite{forster2015imu, leutenegger2015keyframe, usenko2016direct, mur2017visual, qin2018vins} use the batch nonlinear optimization (also known as Bundle Adjustment, BA) to directly minimize the IMU preintegration errors and feature/photometric reprojection errors.
Therefore, it can achieve higher accuracy compared with filter-based approaches.

Although significant progress has been achieved in monocular VIO studies, most methods assume that the measurements of camera and IMU are precisely synchronized without temporal misalignment, and some of them also require that the spatial parameters remain constant and are prerequisite.
However, these conditions are not easily satisfied in practical applications.
In some cases like low-cost and self-assembled devices, accurate factory calibration and hardware synchronization are not available. Besides, the spatial parameters may also drift over time due to wear, tear, sensor reposition, or significant external mechanical stress.

To solve the problems mentioned above, a solution is to perform online initialization and self-calibration. Previously, several online methods for monocular VIO have been developed.
For online spatial calibration, Kelly \textit{et. al} \cite{kelly2011visual} proposed a self-calibration method based on the unscented Kalman filter. The method showed that the full observability of spatial parameters required the sensor suite to undergo both the rotation and acceleration at least two IMU axes.
Li \textit{et. al}\cite{li2013high} proposed a real-time EKF-based VIO algorithm to online calibrate the spatial parameters.
Yang and Shen \cite{yang2017monocular} calibrated the spatial parameters and the initial values (except for IMU bias) with an optimization-based linear estimator. In their extended monocular visual-inertial navigation system (VINS-Mono) \cite{qin2018vins}, the IMU bias is included in the sliding window nonlinear estimator.

For both spatial and temporal online calibration, Li \textit{et al.} \cite{li2014online} treated the time offset as an additional state variable to be estimated along with IMU pose, velocity, biases, feature positions, and extrinsic spatial parameters.
Eckenhoff \textit{et al.} \cite{eckenhoff2019multi} interpolated the IMU poses at an arbitrary intermediate time for all cameras, thus could calibrate the extrinsic spatial and temporal parameters for a multi-camera visual-inertial navigation system.
The observability of spatial-temporal parameters was analyzed by Yang and Huang \textit{et. al} \cite{yang2019degenerate}. Their work showed that the parameters were observable if the sensor platform underwent random motion, and it also identified four degenerate motions that harmed the calibration accuracy.
Although good results have been achieved in the three works mentioned above, they only suit for filter-based methods since they are built upon the multi-state constraint Kalman filter (MSCKF\cite{mourikis2007multi}) framework.

For the optimization-based framework,
Ling \textit{et al.}\cite{ling2018modeling} presented a time-varying model for estimating the camera-IMU time offset using a nonlinear optimization algorithm. This approach can handle the rolling-shutter effects and imperfect sensor synchronization.
Qin \textit{et al.}\cite{qin2018online} recently treated the time offset as a vision factor, and online calibrated it along with features, IMU and camera states in an optimization-based VIO framework.
Nevertheless, the spatial parameters were not considered in these two works.
In \cite{feng2019online}, Feng \textit{et al.} proposed an online spatial-temporal calibration method for monocular direct VIO. Firstly, it estimated the extrinsic rotation and time offset by minimizing the quaternion rotation difference between camera and IMU. Then, a loosely coupled approach introduced in \cite{qin2017robust} was used to recover the initial values. Finally, it proposed a nonlinear optimization algorithm to minimize photometric errors and IMU errors.
Feng's work is similar to ours.
However, one shortcoming of his work was that the extrinsic translation was not initialized. Besides, the adopted loosely coupled approach did not consider the effect of time offset, which might provide rather inaccurate initial estimates for the nonlinear optimization. The noise robustness was also not provided by his work.
Compared with Feng's work, our algorithm outperforms in terms of accuracy and robustness. This is because the short-term motion interpolation algorithms for camera and IMU are both designed, therefore we can consider the effect of time offset and extrinsic translation parameter throughout the three-step process. Furthermore, all the parameters can be continuously optimized by our nonlinear optimization-based algorithm.

The proposed algorithm is an extension of our earlier work \cite{huang2018online}, in which an online initialization method was developed to automatically estimate the initial values and calibrate the camera-IMU transformation for monocular VI-SLAM.
In this work, we extend the previous work to the sensor asynchronous case, by modeling the temporal misalignment between different sensor streams into two short-term motion interpolation algorithms. 

\section{Preliminary}
\label{sec:Preliminary}
This section discusses the IMU model and the preintegration theory. The frame and notation are briefly denoted as follows.
$(\cdot)^w$, $(\cdot)^c$, and $(\cdot)^b$ are respectively the global frame, the local camera frame, and the local IMU body frame.
$\mathbf{T}^w_c=[\mathbf{R}^w_c | s \! \cdot \! \mathbf{p}^w_c]$ is the camera pose in the global frame, where $\mathbf{R}^w_c \in \mathrm{SO}(3)$ and $\mathbf{p}^w_c \in \mathbb{R}^3$ are respectively the camera rotation and position.
On the bootstrapping stage, the camera pose is estimated by a pure monocular VO that subjects to the scale ambiguous problem. Therefore, an unknown visual metric scale $s$ is taken into account.
$\mathbf{T}^w_b = [\mathbf{R}^w_b | \mathbf{p}^w_b]$ is the IMU body pose.
$ \mathbf{T}^b_c = [\mathbf{R}^b_c | \mathbf{p}^b_c]$ is the relative transformation between the camera and IMU, i.e., the extrinsic spatial parameter that should be calibrated.
In the following sections, we also use the inverse representation, i.e., $ \mathbf{T}^c_b = [\mathbf{R}^c_b | \mathbf{p}^c_b]$, for convenience.

\subsection{IMU Model}
\label{subsec:IMU Model}
In principle, given an initial pose and velocity, the IMU pose can be estimated by integrating gyroscope outputs $\omega_b$ and accelerometer outputs $\mathbf{a}_b$. However, the outputs are subject to white sensor noises $\eta_g$ and $\eta_a$ (normally assumed as Gaussian noise), and slow time-varying biases $\mathbf{b}_{g}$ and $\mathbf{b}_{a}$.
The gravitational acceleration $\mathbf{g}^w$ should also be subtracted since it often dominates other measured accelerations.
Thus, the IMU measurement model can be formulated as:
\begin{equation}\label{IMU_measurement_model}
\begin{aligned}
    \omega_b &= \bar{\omega}_b + \mathbf{b}_{g} + \eta_{g}, \\[2mm]
    \mathbf{a}_b &= {\mathbf{R}^w_b}^T  (\bar{\mathbf{a}}_w - \mathbf{g}^w) + \mathbf{b}_{a}  + \eta_{a},
\end{aligned}
\end{equation}
where $\bar{\omega}_b$ and $\bar{\mathbf{a}}_w$ are respectively the angular velocity and linear acceleration that represent the physical dynamic motion properties of the sensor suite in the global frame.

To describe the evolutions of the pose and velocity of IMU body frame, the following kinematic model\cite{Jimenez2009A}\cite{murray2017mathematical} is employed:
\begin{equation}\label{IMU_kinematic_model}
    \dot{\mathbf{R}}^w_b = \mathbf{R}^w_b \cdot {\bar{\omega}_b}^\wedge, \ \  \dot{\mathbf{v}}^w_b = \bar{\mathbf{a}}_w, \ \ \dot{\mathbf{p}}^w_b = \mathbf{v}^w_b.
\end{equation}
Here, $(\cdot)^\wedge$ is the \emph{hat} operator that maps a vector in $\mathbb{R}^3$ to a skew-symmetric matrix.
A property of skew-symmetric matrices that will be used is: Given two vectors $\mathbf{a}, \mathbf{b} \in \mathbb{R}^3$, the cross-product can be expressed as $\mathbf{a} \times \mathbf{b} = \mathbf{a}^\wedge \cdot \mathbf{b} = - \mathbf{b}^\wedge \cdot \mathbf{a}$.

\subsection{IMU Preintegration}
\label{subsec:IMU Preintegration}
Since cameras and IMUs run at different rates, we need a preintegration process to match the IMU measurements with camera frames\cite{forster2017manifold}.
Considering two camera frames captured at $i$ and $j$ ($j > i$) instants, the relationships of IMU rotation $\mathbf{R}^w_b$, velocity $\mathbf{v}^w_b$, and position $\mathbf{p}^w_b$ between the two instants can be given as:
\begin{equation}\label{IMU_summarized}
\begin{aligned}
    \mathbf{R}^w_{b_j} &= \mathbf{R}^w_{b_i} \prod_{k=i}^{j-1} \mathrm{Exp}\left( \left( \omega_{b_k} \! - \! \mathbf{b}_{g_k} \! - \! \eta_{g_k} \right) \Delta t \right),   \\
    \mathbf{v}^w_{b_j} &=  \mathbf{v}^w_{b_i} +  \mathbf{g}^w \Delta t_{ij} +  \sum_{k=i}^{j-1} \mathbf{R}^w_{b_k} \left(\mathbf{a}_{b_k} \! - \! \mathbf{b}_{a_k} \! - \! \eta_{a_k} \right) \Delta t,  \\
    \mathbf{p}^w_{b_j} &=  \mathbf{p}^w_{b_i} \! + \! \sum_{k=i}^{j-1} \!\! \left( \! \mathbf{v}^w_{b_k} \Delta t \! + \! \frac{1}{2} \!\! \left( \mathbf{R}^w_{b_k}(\mathbf{a}_{b_k} \! \! - \! \mathbf{b}_{a_k} \! \! - \! \eta_{a_k}  ) \! + \! \mathbf{g}^w  \right) \! \Delta t^2 \!\! \right),
\end{aligned}
\end{equation}
where $\Delta t$ is the IMU sampling interval, and $\Delta t_{ij} \doteq \sum_{k=i}^{j-1} \Delta t$.
$\mathrm{Exp}(\cdot)$ is the ``vectorized'' version of \emph{exponential map} that transforms a vector $\phi \in \mathfrak{so}(3)$ to a rotation matrix $\mathbf{R} \in \mathrm{SO}(3)$, with $\mathbf{R} = \mathrm{Exp}(\phi) = \mathrm{exp}(\phi^\wedge)$.

Ignoring the measurement noises and assuming the biases remain constant during the preintegration period, the small bias corrections $\delta \mathbf{b}_{g_i}$ and $\delta \mathbf{b}_{a_i}$ could be taken into account to correct the preintegrated terms. Therefore, the expressions in (\ref{IMU_summarized}) can be rewritten as:
\begin{equation}\label{IMU_preintegrate}
\begin{aligned}
    \mathbf{R}^w_{b_j} &= \mathbf{R}^w_{b_i}  \Delta \bar{\mathbf{R}}_{ij} \mathrm{Exp}\left( \mathbf{J}_{\Delta \bar{\mathbf{R}}_{ij}}^g \delta \mathbf{b}_{g_i} \right), \\
    \mathbf{v}^w_{b_j} &=  \mathbf{v}^w_{b_i} \! + \! \mathbf{g}^w \Delta t_{ij} \! + \! \mathbf{R}^w_{b_i} \! \left( \! \Delta \bar{\mathbf{v}}_{ij} \! + \mathbf{J}_{\Delta \bar{\mathbf{v}}_{ij}}^g \delta \mathbf{b}_{g_i} \! + \mathbf{J}_{\Delta \bar{\mathbf{v}}_{ij}}^a \delta \mathbf{b}_{a_i} \! \right) \!,  \\
    \mathbf{p}^w_{b_j} &=  \mathbf{p}^w_{b_i} \! + \! \mathbf{v}^w_{b_i} \Delta t_{ij} \! + \!  \frac{1}{2} \mathbf{g}^w \Delta t_{ij}^2 \\
    & \ \ \ +  \mathbf{R}^w_{b_i} \left( \Delta \bar{\mathbf{p}}_{ij} \!+\! \mathbf{J}_{\Delta \bar{\mathbf{p}}_{ij}}^{g} \delta \mathbf{b}_{g_i} \!+\! \mathbf{J}_{\Delta \bar{\mathbf{p}}_{ij}}^{a} \delta \mathbf{b}_{a_i}  \right),
\end{aligned}
\end{equation}
where the Jacobians $\mathbf{J}^g_{(\cdot)}$ and $\mathbf{J}^a_{(\cdot)}$ indicate how the measurements change due to a change in the bias estimation. The details of the Jacobians can be found in \cite{forster2017manifold}.
The preintegrated terms  $\Delta \bar{\mathbf{R}}_{ij}$, $\Delta \bar{\mathbf{v}}_{ij}$, and $\Delta \bar{\mathbf{p}}_{ij}$ are independent of the states at time $i$ and the gravity.
Given the biases as $\bar{\mathbf{b}}_{g_i}$ and $\bar{\mathbf{b}}_{a_i}$, they can be computed directly from the IMU measured values:
\begin{equation}\label{IMU_dR_dv_dp}
\begin{aligned}
    \Delta \bar{\mathbf{R}}_{ij} &= \prod_{k=i}^{j-1} \mathrm{Exp}\left( \left(\omega_{b_k} - \bar{\mathbf{b}}_{g_i} \right) \Delta t \right), \\
    \Delta \bar{\mathbf{v}}_{ij} &= \sum_{k=i}^{j-1} {\Delta \bar{\mathbf{R}}_{ik} \left( \mathbf{a}_{b_k} - \bar{\mathbf{b}}_{a_i} \right) \Delta t }, \\
    \Delta \bar{\mathbf{p}}_{ij} &= \sum_{k=i}^{j-1}\left( \Delta \bar{\mathbf{v}}_{ik} \Delta t + \frac{1}{2} \Delta \bar{\mathbf{R}}_{ik} \left(\mathbf{a}_{b_k} - \bar{\mathbf{b}}_{a_i} \right) \Delta t^2  \right).
\end{aligned}
\end{equation}

\begin{figure}[htbp]
	\centering
	\includegraphics[width=0.49\textwidth]{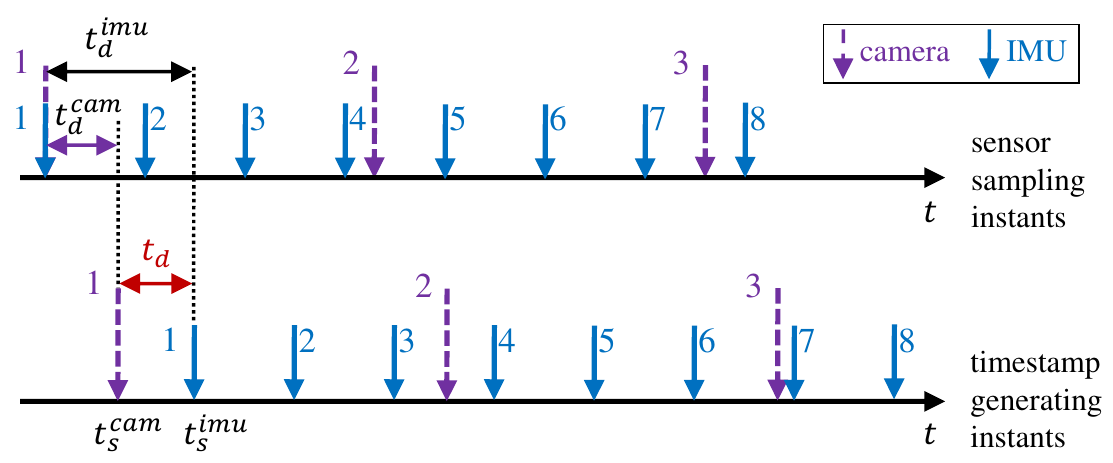}
    \vspace{-1.5em}
	\caption{An example of temporal misalignment between the camera and IMU measurement streams. The upper plot represents the sensor sampling instants. The lower plot shows the timestamp generating instants. The timestamped streams are essentially the sensor measurements that can be obtained.
Here, $t^{imu}_d$ and $t^{cam}_d$ are, respectively, the latency of IMU and camera. $t_d = t^{imu}_d - t^{cam}_d$ is the time offset between the two timestamped streams.
In this case, these two timestamped streams can be aligned by shifting the camera streams with $t_d$ offset or shifting the IMU streams with $-t_d$ offset.
}
	\vspace{-1.0em}
	\label{fig:time_axis}
\end{figure}

\section{Short-term Sensor Motion Interpolation}
\label{sec:short-term sensor motion interpolation}
In this section, we first model the time offset. Then the proposed short-term motion interpolations for camera and IMU are introduced. Finally, we give the pose relationship between the camera and IMU at any timestamp.

\subsection{Time Offset}
\label{subsec:time offset}
In our system, we consider a sensor suite comprising a single camera and a rigidly attached IMU. As shown in Fig. \ref{fig:time_axis}, each of the sensors provides discrete samplings in a constant frequency. However, due to the unsynchronized clocks, transmission delays, sensor response, and operating system overhead, there always exist a latency that makes the measurement (i.e., timestamped) streams misalign with the sampling streams.
Considering the IMU and the camera measurements sampled at the same instant $t$, their timestamps $t^{imu}_s$ and $t^{cam}_s$ are:
\begin{equation}\label{time_relationship_of_sample_and_timestamp}
t^{imu}_s = t + t^{imu}_d, \\ \ \ \ \ t^{cam}_s = t + t^{cam}_d,
\end{equation}
where $t^{imu}_d$ and $t^{cam}_d$ are respectively the latency of IMU and camera. Therefore, the unknown time offset (i.e., the temporal parameter) $t_d$ can be defined as follows:
\begin{equation}\label{time_offset_definition}
  t_d \doteq t^{imu}_d - t^{cam}_d = t^{imu}_s - t^{cam}_s.
\end{equation}

It is worth noting that the time offset is identifiable, while the individual latencies of the sensors are indistinguishable unless additional state information is available\cite{li2014online}.
According to (\ref{time_relationship_of_sample_and_timestamp}) and (\ref{time_offset_definition}), the IMU and camera measurement streams can be aligned by shifting the camera streams with $t_d$ offset or shifting the IMU streams with $-t_d$ offset, which results in the following pose relationships:
\begin{align}
\mathbf{T}^w_b(t) = \mathbf{T}^w_c(t+t_d) \cdot \mathbf{T}^c_b,
\label{pose_relationship_camera_to_IMU} \\
\mathbf{T}^w_c(t) = \mathbf{T}^w_b(t-t_d) \cdot \mathbf{T}^b_c.
\label{pose_relationship_IMU_to_camera}
\end{align}
Here, $\mathbf{T}^w_b(t)$ and $\mathbf{T}^w_c(t)$ are, respectively, the pose of IMU and camera at timestamp $t$.
The formula (\ref{pose_relationship_camera_to_IMU}) indicates that an IMU measurement with timestamp $t$ is aligned with the camera measurement with timestamp $t+t_d$, while the formula (\ref{pose_relationship_IMU_to_camera}) indicates that a camera measurement with timestamp $t$ is aligned with the IMU measurement with timestamp $t-t_d$.


\subsection{Motion Interpolation}
\label{subsec:motion interpolation}
In the following, we use simplified notations for convenience of expression. For example, we denote the camera rotation at $t_i$ in the world frame as $\mathbf{R}^w_{c_i}$, which is a simplification of $\mathbf{R}^w_c(t_i)$.
We denote the camera pose at $t_i+t_d$ as $\mathbf{T}^w_{c_i,t_d} = (\mathbf{R}^w_{c_i,t_d} | s \cdot \mathbf{p}^w_{c_i,t_d})$, which is a simplification of $\mathbf{T}^w_c(t_i+t_d)$.
Without further explanation, the similar simplification is applied to other notations.

\subsubsection{Camera motion interpolation}
\label{subsecsec:Camera motion interpolation}
By assuming the camera moves in constant angular and linear velocities in a short period, the camera pose at an arbitrary time can be interpolated with its nearest camera pose, angular velocity, and linear velocity.
Considering two camera poses $\mathbf{T}^w_{c_i}$ and $\mathbf{T}^w_{c_j}$ that estimated by monocular VO at timestamp $t_i$ and $t_j$, the camera angular velocity $\omega_{c_i}$ and linear velocity $\tilde{\mathbf{v}}_{c_i}$ at $t_i$ can be approximated, as follows:
\begin{equation}\label{camera_angular_and_linear_velocity}
\begin{split}
\omega_{c_i} &\approx \mathrm{Log}({\mathbf{R}^w_{c_i}}^T \mathbf{R}^w_{c_j}) / (t_j - t_i), \\
\tilde{\mathbf{v}}_{c_i} &\approx (\mathbf{p}^w_{c_j} - \mathbf{p}^w_{c_i}) / (t_j - t_i),
\end{split}
\end{equation}
where $\mathrm{Log}(\cdot)$ is the ``vectorized'' version of \emph{logarithm map} that transforms a rotation matrix $\mathbf{R} \neq \mathbf{I}$ to a vector $\phi$, with $\phi = \mathrm{Log}(\mathbf{R}) = \mathrm{ln}(\mathbf{R})^\vee$.
Here, $(\cdot)^\vee$ is the \emph{vee} operator that maps a skew-symmetric matrix in $\mathbb{R}^{3\times3}$ to a vector in $\mathbb{R}^3$.
Note that the velocity term $\tilde{\mathbf{v}}_{c_i}$  is not the actual camera linear velocity in the global frame but subjects to an unknown metric scale.

According to (\ref{camera_angular_and_linear_velocity}), the camera rotation and position at time $t_i+t_d$ can be interpolated as follows:
\begin{equation}\label{camera_rotation_and_position_interpolation}
\begin{split}
\mathbf{R}^w_{c_i,t_d} &\approx \mathbf{R}^w_{c_i} \mathrm{Exp}(\omega_{c_i}  t_d) , \\
\mathbf{p}^w_{c_i,t_d} &\approx \mathbf{p}^w_{c_i} + \tilde{\mathbf{v}}_{c_i}  t_d.
\end{split}
\end{equation}

\subsubsection{IMU motion interpolation}
\label{subsubsec:IMU motion interpolation}
Similarly, the IMU rotation and position at time $t_i-t_d$ can be interpolated as:
\begin{equation}\label{IMU_rotation_and_position_interpolation}
\begin{split}
\mathbf{R}^w_{b_i,-t_d} &\approx \mathbf{R}^w_{b_i}  \mathrm{Exp}(-\bar{\omega}_{b_i}  t_d) , \\
\mathbf{p}^w_{b_i,-t_d} &\approx \mathbf{p}^w_{b_i} - \mathbf{v}^w_{b_i}  t_d,
\end{split}
\end{equation}
where $\bar{\omega}_{b_i}$ is the actual IMU body angular velocity at timestamp $t_i$. $\mathbf{v}^w_{b_i}$ is the IMU body linear velocity expressed in the global frame.

\subsection{Transformation Relationship}
\label{subsec:transformation relationship}
Considering the time offset $t_d$ and the metric scale $s$, the rotation and position of IMU body at timestamp $t_i$ can be derived from camera pose according to (\ref{pose_relationship_camera_to_IMU}), as follows:
\begin{align}
\mathbf{R}^w_{b_i} &= \mathbf{R}^w_{c_i,t_d}  \mathbf{R}^c_b \approx \mathbf{R}^w_{c_i}  \mathrm{Exp}(\omega_{c_i} t_d)  \mathbf{R}^c_b, \label{pose_relationship_camera_to_IMU_R} \\
\begin{split}
\mathbf{p}^w_{b_i} &=  \mathbf{R}^w_{c_i,t_d}  \mathbf{p}^c_b + s \cdot \mathbf{p}^w_{c_i,t_d} \\
&\approx \mathbf{R}^w_{c_i}  \mathrm{Exp}(\omega_{c_i} t_d)  \mathbf{p}^c_b + s \cdot ( \mathbf{p}^w_{c_i} + \tilde{\mathbf{v}}_{c_i}  t_d ).
\end{split}\label{pose_relationship_camera_to_IMU_p}
\end{align}

Similarly, the camera pose at timestamp $t_i$ can be derived from IMU pose according to (\ref{pose_relationship_IMU_to_camera}), as follows:
\begin{align}
\mathbf{R}^w_{c_i} &= \mathbf{R}^w_{b_i,-t_d}  \mathbf{R}^b_c \approx \mathbf{R}^w_{b_i}  \mathrm{Exp}(-\bar{\omega}_{b_i} t_d)  \mathbf{R}^b_c, \label{pose_relationship_IMU_to_camera_R} \\
\begin{split}
\mathbf{p}^w_{c_i} &= \mathbf{R}^w_{b_i,-t_d}  \mathbf{p}^b_c + \mathbf{p}^w_{b_i,-t_d} \\
&\approx \mathbf{R}^w_{b_i}  \mathrm{Exp}(-\bar{\omega}_{b_i} t_d)  \mathbf{p}^b_c + \mathbf{p}^w_{b_i} - \mathbf{v}^w_{b_i} t_d,
\end{split}\label{pose_relationship_IMU_to_camera_p}
\end{align}
where the metric scale term is eliminated in (\ref{pose_relationship_IMU_to_camera_p}) since it is observable by IMU integration.
In the following sections, the formulae (\ref{pose_relationship_camera_to_IMU_R}) and (\ref{pose_relationship_camera_to_IMU_p}) are used to derive the three-step process.
The formulae (\ref{pose_relationship_IMU_to_camera_R}) and (\ref{pose_relationship_IMU_to_camera_p}) are used to derive the feature reprojection error.


\begin{figure}[tbp]
	\centering
	\includegraphics[width=0.45\textwidth]{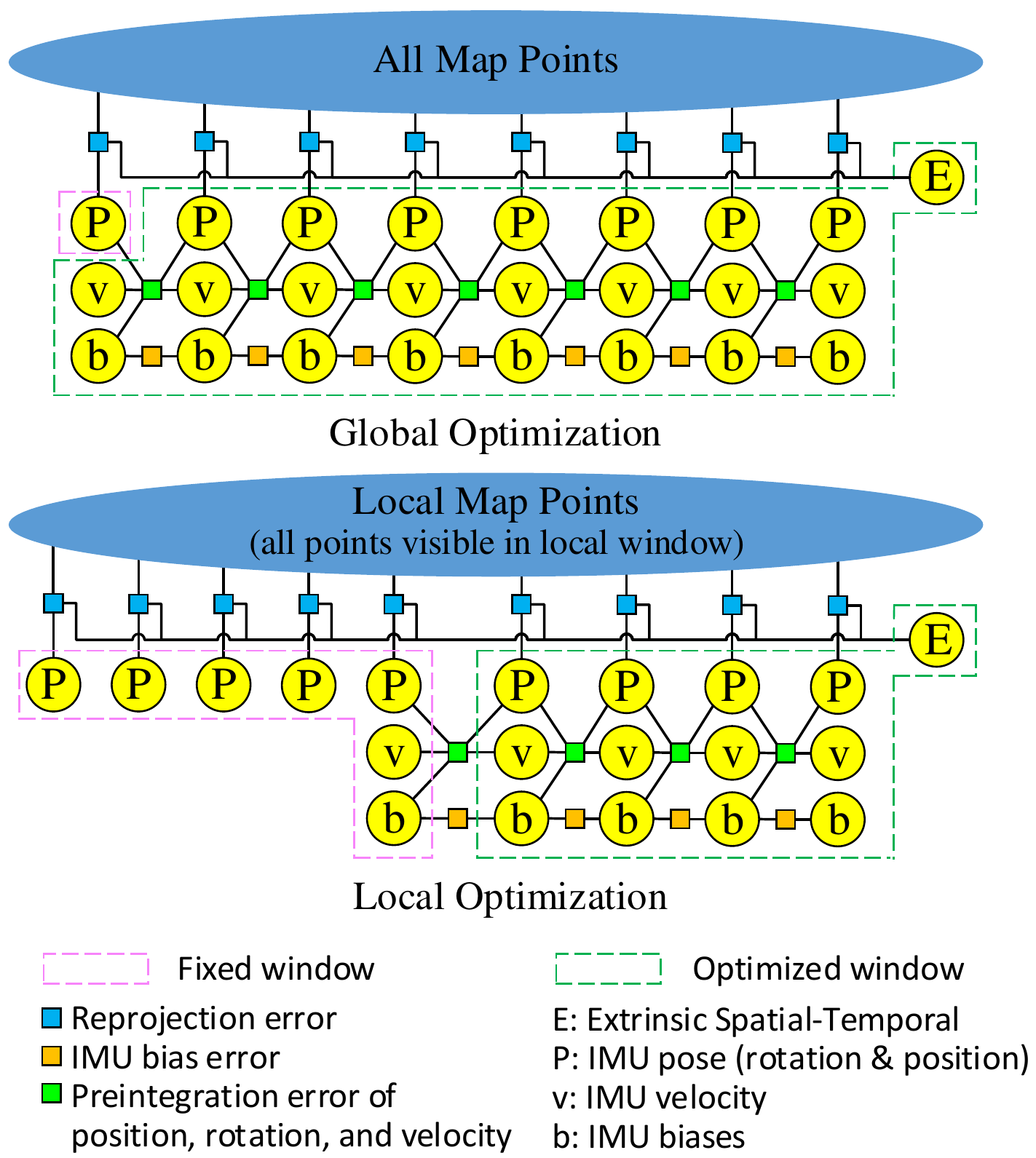}
	\caption{Comparison between global optimization (top) and local optimization (bottom). The variables in the optimized window are optimized during optimization, while the variables in the fixed window remain constant.}
	\label{fig:global_local_BA}
\end{figure}

\section{Visual-Inertial State Estimation}
\label{sec:Visual-Inertial_State_Estimation}
In this section, the states of the studied system and the proposed visual-inertial nonlinear optimization are discussed. The feature reprojection error and IMU preintegration error are also introduced.

\subsection{System States}
\label{subsec:system_state}
In our method, the state vector for the $i$th keyframe is defined as follows:
\begin{equation}\label{state_for_each_keyframe}
  \mathbf{x}_i = [\mathbf{R}^w_{b_i}, \mathbf{p}^w_{b_i}, \mathbf{v}^w_{b_i}, \mathbf{b}_{g_i}, \mathbf{b}_{a_i}, \mathbf{p}^w_{i1}, \mathbf{p}^w_{i2}, ..., \mathbf{p}^w_{im} ],
\end{equation}
where $\mathbf{p}^w_{ik} \in \mathbb{R}^3$ is the $k$th map point observed by the $i$th keyframe.
The full states of a nonlinear optimization are defined as follows:
\begin{equation}\label{full_states}
  \mathcal{X} = [\mathbf{x}_l, \mathbf{x}_{l+1}, ..., \mathbf{x}_n, \mathbf{R}^b_c, \mathbf{p}^b_c, t_d],
\end{equation}
where $n$ and $l$ are respectively the newest and oldest keyframe indexes of an optimized window with size of $\mathcal{L}$.
As shown in Fig. \ref{fig:global_local_BA}, the setting of $l$ depends on the type of optimization.
For global optimization, except for the position and rotation of the first keyframe as it is usually regarded as the world reference frame, all the other states are optimized. Therefore, the $l$ is set to $1$.
For local optimization, we optimize the extrinsic spatial-temporal parameters and the states of the keyframes that contained in a local window. As a result, the $l$ is set to $n-\mathcal{L}+1$.
The retractions of the system states are provided in Appendix \ref{appendix: State Update}.

\subsection{Nonlinear Optimization With Time Offset}
\label{subsec:nonlinear_optimization}
In the nonlinear optimization, both the IMU preintegration error and the feature reprojection error are minimized, as follows:
\begin{equation}\label{nonlinear_optimization}
\mathcal{X}^* = \mathop{\arg\min} \limits_{\mathcal{X}}  \sum_{i = l}^{n} \! \left( \sum_{k}{\mathbf{E}_{proj}(k, i)} + \mathbf{E}_{imu} (i-1, i) \right).
\end{equation}
Here, $\mathbf{E}_{proj}(k, i)$ is the feature reprojection error term for a given matched $k$th map point that observed by the $i$th keyframe. $\mathbf{E}_{imu} (i-1, i)$ is the IMU preintegration error term that links keyframe $i$ and its previous keyframe $i\!-\!1$.

\subsubsection{Feature reprojection error}
Considering a 3D map point $\mathbf{p}^w_k \in \mathbb{R}^3$ in the global frame that is observed by the $i$th keyframe and matched to a 2D image feature, the map point can be transformed into the local camera frame using (\ref{pose_relationship_IMU_to_camera_R}) and (\ref{pose_relationship_IMU_to_camera_p}), as follows:
\begin{equation}\label{p_ci_k}
\mathbf{p}^{c_i}_k = \mathbf{R}^c_b   \mathrm{Exp}(\tilde{\omega}_{b_i}  t_d) {\mathbf{R}^w_{b_i}}^T  \left( \mathbf{p}^w_k  - \mathbf{p}^w_{b_i} + \mathbf{v}^w_{b_i} t_d \right) + \mathbf{p}^c_b,
\end{equation}
where we use $\tilde{\omega}_{b_i} = \omega_{b_i} - \bar{\mathbf{b}}_{g_i} - \delta \mathbf{b}_{g_i} $ to approximate $\bar{\omega}_{b_i}$ by ignoring the white sensor noise.
Here, $\bar{\mathbf{b}}_{g_i}$ is the assumed constant gyroscope bias during IMU preintegration, and $\delta \mathbf{b}_{g_i}$ is the bias correction.

As a result, the feature reprojection error can be defined as follows:
\begin{equation}\label{reprojection_error}
\mathbf{E}_{proj}{(k,i)} = \rho \left(\left(\mathbf{u}^i_k - \pi(\mathbf{p}^{c_i}_k) \right)^T \Sigma_{k} \left( \mathbf{u}^i_k - \pi(\mathbf{p}^{c_i}_k) \right)  \right),
\end{equation}
where $\pi:\mathbb{R}^3\rightarrow\Omega$ is the projection function of pinhole camera model\cite{Hartley2004}, which transforms a 3D point  in the camera frame into a 2D point on the image plane.
$\mathbf{u}^i_k \in \mathbb{R}^2$ is the pixel location of the matched feature.
$\Sigma_{k}$ is the information matrix associated with the feature scale.
$\rho$ is a Huber robust cost function.

Note that the formula (\ref{reprojection_error}) constraints the extrinsic spatial-temporal parameters, as well as the IMU pose, IMU velocity, and map points. The Jacobians of reprojection error w.r.t. the states are derived in Appendix \ref{appendix: Jacobians of Reprojection Errors}.

\subsubsection{IMU preintegration error} With a slight abuse of notation, here we adopt $i$ and $j$ to denote two consecutive keyframes for convenient. The IMU preintegration error term $\mathbf{E}_{imu}(i,j)$ is defined as:
\begin{equation}\label{IMU_error}
\begin{aligned}
    \mathbf{E}_{imu}{(i,j)} &= \rho \left( [\mathbf{e}_R^T \ \mathbf{e}_v^T \ \mathbf{e}_p^T] \Sigma_I [\mathbf{e}_R^T \ \mathbf{e}_v^T \ \mathbf{e}_p^T]^T \right) + \rho \left( \mathbf{e}_b^T \Sigma_R \mathbf{e}_b \right), \\[2mm]
    \mathbf{e}_R &=  \mathrm{Log} \left( \big(\Delta \bar{\mathbf{R}}_{i,j} \mathrm{Exp}(\mathbf{J}_{\Delta \bar{\mathbf{R}}_{i,j}}^{g} \delta\mathbf{b}_{g}^i) \big)^T {\mathbf{R}^w_{b_i}}^T \mathbf{R}^w_{b_j}  \right),  \\
    \mathbf{e}_v  &= {\mathbf{R}^w_{b_i}}^T \left( \mathbf{v}^w_{b_j} - \mathbf{v}^w_{b_i} - \mathbf{g}^w \Delta t_{ij}  \right) \\
    & \ \ \  - \left( \Delta \bar{\mathbf{v}}_{ij} + \mathbf{J}_{\Delta \bar{\mathbf{v}}_{ij}}^{g} \delta\mathbf{b}_{g}^i + \mathbf{J}_{\Delta \bar{\mathbf{v}}_{ij}}^{a} \delta\mathbf{b}_{a}^i \right), \\
    \mathbf{e}_p  &= {\mathbf{R}^w_{b_i}}^T \left(  \mathbf{p}^w_{b_j} -  \mathbf{p}^w_{b_i} - \mathbf{v}^w_{b_i} \Delta t_{ij} - \frac{1}{2} \mathbf{g}^w \Delta t_{ij}^2 \right)   \\
    & \ \ \ - \left( \Delta \bar{\mathbf{p}}_{ij} + \mathbf{J}_{\Delta \bar{\mathbf{p}}_{ij}}^{g}  \delta\mathbf{b}_{g}^i + \mathbf{J}_{\Delta \bar{\mathbf{p}}_{ij}}^{a} \delta\mathbf{b}_{a}^i  \right), \\[1mm]
    \mathbf{e}_b &= \delta \mathbf{b}^j - \delta \mathbf{b}^i,
\end{aligned}
\end{equation}
where $\delta\mathbf{b}^j = [{\delta\mathbf{b}_{g}^j}^T \ {\delta\mathbf{b}_{a}^j}^T ]^T$.
$\mathbf{e}_R$, $\mathbf{e}_v$, and $\mathbf{e}_p$ are respectively the errors of the integrated rotation, velocity, and position. $\mathbf{e}_b$ is the bias errors at $i$ and $j$ time instants.
$\Sigma_I$ and $\Sigma_R$ are the information matrices of the preintegration and the bias random walk, respectively.

\section{Online Initialization and Extrinsic Spatial-Temporal Calibration}
\label{sec:IMU Initialization and Extrinsic Spatial-Temporal Calibration}
This section details the proposed three-step process to jointly calibrate the extrinsic spatial and temporal parameters between camera and IMU, as well as to estimate the initial values of velocity, scale, gravity, and IMU biases. To make all variables observable, our method requires the monocular visual odometry front-end to have been run for a few seconds to collect several keyframes. The pose and feature measurements estimated by the front-end are subject to an unknown metric scale.
When a new keyframe is collected, this three-step process will be performed once.

\subsection{Step-1: Estimating Gyroscope Bias, and Calibrating Extrinsic Rotation and Time Offset}
\label{subsec:Gyroscope Bias Estimation, Extrinsic Rotation and Temporal Calibration}
%
%

In our previous work \cite{huang2018online}, the gyroscope bias and the extrinsic rotation could be separately estimated using an iterative strategy. However, in this work, it is difficult to iteratively estimate the time offset and the extrinsic rotation, since they are tightly coupled. Instead, we directly estimate all these parameters in a minimum function. The derivation process is introduced in the following.

In the initialization stage, the gyroscope bias is assumed as a constant value as it changes slowly over time.
The rotation relationships of two consecutive keyframes at timestamp $i$ and $i+1$ can be described as:
\begin{equation}\label{rotation_relationship}
\mathbf{R}^w_{b_i} =  \mathbf{R}^w_{c_i,t_d}  \mathbf{R}^c_b, \\
\ \ \ \mathbf{R}^w_{b_{i+1}} =  \mathbf{R}^w_{c_{i+1}, t_d}  \mathbf{R}^c_b.
\end{equation}
Here, $\mathbf{R}^w_{b_i}$ and $\mathbf{R}^w_{b_{i+1}}$ are the IMU body rotations that derived by transforming the interpolated camera rotations
By substituting (\ref{IMU_preintegrate}) into (\ref{rotation_relationship}), the difference between the preintegrated rotation and the transformed results are:
\begin{equation}\label{e_Rbc_td_gyroBias}
\begin{aligned}
	\mathbf{e}_{rot_{i,i+1}} &= \mathrm{Log} \left( \left(\Delta \bar{\mathbf{R}}_{i,i+1} \mathrm{Exp}\left(\mathbf{J}_{\Delta \bar{\mathbf{R}}}^g \delta \mathbf{b}_g\right) \right)^T \!  \mathbf{R}^b_c  \right.  \\[1mm]
	&  \left. \cdot \mathrm{Exp}\left(-\omega_{c_i} t_d\right) \mathbf{R}^{c_i}_w\mathbf{R}^w_{c_{i+1}} \mathrm{Exp}\left(\omega_{c_{i+1}}  t_d \right)  \mathbf{R}^c_b \right),
\end{aligned}
\end{equation}
where $\mathbf{R}^c_b$ and $\mathbf{R}^{c_i}_w$ are, respectively,                                           the inverse of $\mathbf{R}^b_c$ and $\mathbf{R}^w_{c_i}$.

Considering there are $N$ keyframes determined by monocular VO front-end, the spatial rotation, time offset, and gyroscope bias can be estimated by minimizing the rotation difference for all keyframes, as follows:
\begin{equation}\label{gyroBias_Rbc_td_min_func}
  \delta b^*_g, t_d^*, {\mathbf{R}^b_c}^* = \mathop{\arg\min} \limits_{\delta b_g, t_d, {\mathbf{R}^b_c}} \sum_{i = 1}^{N-1} \|\mathbf{e}_{rot_{i,i+1}} \|_{\Sigma_{\Delta \! \mathbf{R}}},
\end{equation}
where $\Sigma_{\Delta \! \mathbf{R}}$ is the information matrices of the preintegrated rotation.
The Jacobians of $\mathbf{e}_{rot_{i,i+1}}$ w.r.t. the optimized states are derived in Appendix \ref{appendix: Jacobians of the first process}.
The preintegration terms are re-computed once we obtain a new gyroscope bias estimation.

\subsection{Step-2: Approximating Scale, Gravity, and Extrinsic Translation}
\label{subsec:Scale_Gravity and Translation Approximation}
Once the extrinsic rotation ${\mathbf{R}^b_c}^*$ and time offset $t_d^*$ have been calibrated, the scale $s$, gravity $\mathbf{g}^w$, and extrinsic translation $\mathbf{p}^c_b$ can be approximately estimated.
Since the accelerometer bias is not considered in this step, $\mathbf{b}_a$, $\mathbf{J}_{\Delta \bar{\mathbf{p}}}^{a}$, and $\mathbf{J}_{\Delta \bar{\mathbf{v}}}^{a}$ are temporarily set to zero.
Also, by re-computing the preintegration terms after the gyroscope bias estimation and assuming the gyroscope bias is constant, $\mathbf{J}_{\Delta \bar{\mathbf{p}}}^{g}$ and $\mathbf{J}_{\Delta \bar{\mathbf{v}}}^{g}$ can be set to zero.

By substituting (\ref{pose_relationship_camera_to_IMU_R}) and (\ref{pose_relationship_camera_to_IMU_p}) into the third equation of (\ref{IMU_preintegrate}), the position relationship between two consecutive keyframes can be obtained:
\begin{equation}\label{trans_calib_two_KFs}
\begin{aligned}
s\cdot \mathbf{p}^w_{c_{i+1},t_d} &= s \cdot \mathbf{p}^w_{c_i,t_d} + \mathbf{v}^w_{b_i} \Delta t_{i,i+1} + \frac{1}{2} \mathbf{g}^w \Delta t_{i,i+1}^2 \\[1mm]
& \!\!\!\!\!\!\!\!\! + \mathbf{R}^w_{c_i,t_d} {\mathbf{R}^c_b}^* \Delta \bar{\mathbf{p}}_{i,i+1} + (\mathbf{R}^w_{c_i,t_d} \! - \! \mathbf{R}^w_{c_{i+1,t_d}} ) \!\cdot\! \mathbf{p}^c_b,
\end{aligned}
\end{equation}
where ${\mathbf{R}^c_b}^*$ is the result estimated in the first step process.
By considering three consecutive keyframes and using the second equation of (\ref{IMU_preintegrate}) to eliminate the velocity term $\mathbf{v}^w_{b_i}$, we have:
\begingroup
\renewcommand*{\arraystretch}{1.0}  
\setlength{\arraycolsep}{2.0pt}     
    \begin{equation}\label{trans_calib_three_KFs}
    \left[ \begin{array}{ccc}
    \lambda(i) & \beta(i) & \varphi(i)
    \end{array}  \right] \\
    \left[ \begin{array}{c}
    s \\
    \mathbf{g}^w \\
    \mathbf{p}^c_b
    \end{array} \right] = \gamma(i) .
    \end{equation}
\endgroup
When writing keyframes $i$, $i+1$, $i+2$ as $1$, $2$, $3$, $\lambda(i)$, $\beta(i)$, $\varphi(i)$, and $\gamma(i)$ can be expressed as: 
\begin{equation}\label{trans_calib_lambda_beta_varphi_gamma}
\begin{aligned}
\lambda(i)  &= ( \mathbf{p}^w_{c_2,t_d} \!\!- \mathbf{p}^w_{c_1,t_d}  ) \Delta t_{23} - ( \mathbf{p}^w_{c_3,t_d} \!\!- \mathbf{p}^w_{c_2,t_d}  ) \Delta t_{12},    \\[1mm]
\beta(i)    &=  \frac{1}{2} ( \Delta t_{12} \Delta t_{23}^2 + \Delta t_{12}^2 \Delta t_{23} ) \mathbf{I}_{3 \times 3},   \\[1mm]
\varphi(i)  &= (\mathbf{R}^w_{c_2,t_d} \!\!- \mathbf{R}^w_{c_3,t_d})\Delta t_{12} \!- (\mathbf{R}^w_{c_1,t_d} \!\!- \mathbf{R}^w_{c_2,t_d})\Delta t_{23},    \\[1mm]
\gamma(i)   &= \mathbf{R}^w_{c_1,t_d} {\mathbf{R}^c_b}^* \left( \Delta \bar{\mathbf{p}}_{12} \Delta t_{23} \!- \Delta \bar{\mathbf{v}}_{12} \Delta t_{12}\Delta t_{23} \right) \\[1mm]
& \ \ \ - \mathbf{R}^w_{c_2,t_d} {\mathbf{R}^c_b}^* \Delta \bar{\mathbf{p}}_{23} \Delta t_{12} .
\end{aligned}
\end{equation}

With $N$ keyframes, we can obtain $N-2$ relations like (\ref{trans_calib_three_KFs}). All relations can be stacked into a linear over-determined equation $\mathbf{B}_{3(N-2)\times7} \cdot \mathbf{x}_{7\times1} = \mathbf{C}_{3(N-2)\times1}$ with weights for outlier handling as described in our previous work \cite{huang2018online}. This equation can be solved via Singular Value Decomposition (SVD) to get the metric scale $s^*$, gravity vector ${\mathbf{g}^w}^*$, and extrinsic translation ${\mathbf{p}^c_b}^*$.
Note that there are $3(N-2)$ equations and 7 unknowns, at least 5 keyframes is required to calculate a solution.

\subsection{Step-3: Estimating Accelerometer Bias, and Refining Scale, Gravity, and Translation}
\label{subsec:Accelerometer Bias Estimation and Scale Gravity and Translation Refinement}
Note that the accelerometer bias and gravity are difficult to distinguish, the accelerometer bias was temporarily set to zero and a rough gravity ${\mathbf{g}^w}^*$ was obtained in the second step. 
In this step, in order to estimate the accelerometer bias and refine the metric scale, gravity, and extrinsic translation, we take the magnitude of gravitational acceleration into account.


Using the already estimated ${\mathbf{g}^w}^*$, the rotation between the earth fixed reference frame $\{e\}$ and the world frame $\{w\}$ can be obtained as:
\begin{equation}\label{refine_Rwi}
    \begin{aligned}
        \mathbf{R}^w_e &= \mathrm{Exp}(\tilde{\mathbf{v}}\theta),    \\[1mm]
        \tilde{\mathbf{v}} &=  \frac{\tilde{\mathbf{g}}^e \times \tilde{\mathbf{g}}^w }{\| \tilde{\mathbf{g}}^e \times \tilde{\mathbf{g}}^w \|}, \ \ \theta = \mathrm{atan2}(\| \tilde{\mathbf{g}}^e \times \tilde{\mathbf{g}}^w \|, \ \tilde{\mathbf{g}}^e \cdot \tilde{\mathbf{g}}^w ), \\[1mm]
        \tilde{\mathbf{g}}^w &= {\mathbf{g}^w}^* / \|{\mathbf{g}^w}^*\| , \ \
        \tilde{\mathbf{g}}^e = \mathbf{G}^e / \|\mathbf{G}^e \|, \ \
        \mathbf{G}^e = \left[ 0 \ 0 \ {-\mathrm{G}} \right]^T,
    \end{aligned}
\end{equation}
where $\tilde{\mathbf{v}}$ and $\theta$ are respectively the rotation axis and the rotation angle.  $\mathbf{G}^e$ is the gravity vector expressed in \{e\}. $\mathrm{G}$ is the magnitude of the gravitational acceleration (normally $\mathrm{G}=9.81 m\cdot s^{-2}$).
This rotation can be optimized by appending a perturbation $\delta \theta \in \mathbb{R}^{3\times1}$, as follows:
\begin{equation}\label{refine_gw}
    \begin{aligned}
        \mathbf{g}^w &= \mathbf{R}^w_e \mathrm{Exp}(\delta \theta)  \cdot  \mathbf{G}^e \ \! {\approx} \ \! \mathbf{R}^w_e \cdot \mathbf{G}^e  -  \mathbf{R}^w_e \cdot {\mathbf{G}^e}^\wedge \cdot  \delta \theta, \!\!\!
    \end{aligned}
\end{equation}
where the first-order approximation of exponential map (see Appendix \ref{appendix: preliminaries}) is applied.
By substituting (\ref{refine_gw}) into (\ref{trans_calib_two_KFs}) and further considering a constant accelerometer bias, we have:
\begin{equation}\label{refine_calib_two_KFs}
\begin{aligned}
    s \!\cdot \mathbf{p}^w_{c_{i+1},t_d} &= s \!\cdot \mathbf{p}^w_{c_{i},t_d} \!+ \mathbf{v}^w_{b_i} \Delta t_{i,i+1} \!- \frac{1}{2} \mathbf{R}^w_e \!\cdot {\mathbf{G}^e}^\wedge\! \cdot  \delta \theta \Delta t_{i,i+1}^2 \\[1mm]
    & \ \ \ + \mathbf{R}^w_{c_{i},t_d} {\mathbf{R}^c_b}^* (\Delta \bar{\mathbf{p}}_{i,i+1} \!+ \mathbf{J}_{\Delta \bar{\mathbf{p}}_{i,i+1}}^{a}\delta \mathbf{b}_{a} ) \\[1mm]
    & \ \ \ + (\mathbf{R}^w_{c_{i},t_d} \!- \mathbf{R}^w_{c_{i+1},t_d}) \cdot \mathbf{p}^c_b + \frac{1}{2} \mathbf{R}^w_e \!\cdot \mathbf{G}^e \Delta t_{i,i+1}^2 .
\end{aligned}
\end{equation}

Similar to (\ref{trans_calib_three_KFs}), the velocity term can be eliminated by considering three consecutive keyframes and using the second equation of (\ref{IMU_preintegrate}), which results in:
\begingroup
\renewcommand*{\arraystretch}{1.0}  
\setlength{\arraycolsep}{2.0pt}     
    \begin{equation}\label{refine_calib_three_KFs}
        \left[ \begin{array}{cccc}
        \lambda(i) & \phi(i) & \zeta(i) & \xi(i)
        \end{array}  \right] \\
        \left[ \begin{array}{c}
        s \\
        \delta \theta_{xy} \\
        \delta \mathbf{b}_{a}    \\
        \mathbf{p}^c_b
        \end{array} \right] \\
        = \psi(i),
    \end{equation}
\endgroup
where $\lambda(i)$ remains the same as in (\ref{trans_calib_lambda_beta_varphi_gamma}), and $\phi(i)$, $\zeta(i)$, $\xi(i)$, and $\psi(i)$ are computed as follows:
\begin{equation}\label{refine_calib_phi_zeta_xi}
\begin{aligned}
\phi(i) &=  \left[ - \frac{1}{2}  \mathbf{R}^w_e \!\cdot\! {\mathbf{G}^e}^\wedge \!\cdot\! ( \Delta t_{12} \Delta t_{23}^2 + \Delta t_{12}^2 \Delta t_{23} ) \right]_{(:,1:2)},   \\[1mm]
\zeta(i)&= \mathbf{R}^w_{c_1,t_d} {\mathbf{R}^c_b}^* \left( \mathbf{J}_{\Delta \bar{\mathbf{v}}_{12}}^{a} \Delta t_{12} \Delta t_{23} - \mathbf{J}_{\Delta \bar{\mathbf{p}}_{12}}^{a} \Delta t_{23} \right) \\[1mm]
& \ \ \ + \mathbf{R}^w_{c_2,t_d} {\mathbf{R}^c_b}^* \mathbf{J}_{\Delta \bar{\mathbf{p}}_{23}}^{a} \Delta t_{12}, \\[1mm]
\xi(i)  &= (\mathbf{R}^w_{c_2,t_d} - \mathbf{R}^w_{c_3,t_d}) \Delta t_{12} - (\mathbf{R}^w_{c_1,t_d} - \mathbf{R}^w_{c_2,t_d}) \Delta t_{23}, \\[1mm]
\psi(i) &= \mathbf{R}^w_{c_1,t_d} {\mathbf{R}^c_b}^* \left( \Delta \bar{\mathbf{p}}_{12} \Delta t_{23} -  \Delta \bar{\mathbf{v}}_{12} \Delta t_{12} \Delta t_{23} \right) \\[1mm]
& \ \ \ - \mathbf{R}^w_{c_2,t_d} {\mathbf{R}^c_b}^* \Delta \bar{\mathbf{p}}_{23} \Delta t_{12} \\[1mm]
& \ \ \ - \frac{1}{2} \mathbf{R}^w_e \cdot \mathbf{G}^e ( \Delta t_{12} \Delta t_{23}^2 + \Delta t_{12}^2 \Delta t_{23} ),
\end{aligned}
\end{equation}
where $[\cdot]_{(:,1:2)}$ means the first two columns of the matrix.
With $N$ keyframes, a linear over-determined equation $\mathbf{D}_{3(N-2)\times 9} \cdot \mathbf{y}_{9\times 1} = \mathbf{E}_{3(N-2)\times 1}$ with weights for outlier handling can be constructed to calculate a solution of $s^*$, $\delta \theta_{xy}^*$, $\delta \mathbf{b}_{a}^*$, and ${\mathbf{p}^c_b}^*$.
Since the accelerometer bias is set to zero when integrating $\Delta \bar{\mathbf{R}}_{i,i+1}$, $\Delta \bar{\mathbf{v}}_{i,i+1}$, and $\Delta \bar{\mathbf{p}}_{i,i+1}$, the final estimated accelerometer bias is $\mathbf{b}_{a}^* = \mathbf{0}_{3\times 1} + \delta \mathbf{b}_{a}^* = \delta \mathbf{b}_{a}^*$. The gravity is refined by appending the perturbation, i.e., ${\mathbf{g}^w}^* = \mathbf{R}^w_e \mathrm{Exp}(\delta \theta^*) \cdot \mathbf{G}^e$.

\subsection{Compensation of Time Offset and Initialization Trick}
\label{subsec:Compensation of Time Offset and Initialization Trick}
After each execution of the three-step process or nonlinear optimization, the time offset is compensated by shifting the timestamps of subsequent visual measurements, i.e., ${t^{cam}_s}' = t^{cam}_s + t_d$. Then, the system estimates a new time increment $\delta t_d$ between the compensated visual measurement and inertial measurement in the following.
Note that we do not have any prior knowledge about the sensor's temporal misalignment, and all the collected keyframes are utilized to perform the three-step process.
If the misalignment is large, the old keyframes whose timestamp does not been compensated by new estimated time offsets will have bad impacts on following executions.
One trick that we adopted to deal with this problem is to discard the old keyframes.
Specifically, if the time interval estimated by (\ref{gyroBias_Rbc_td_min_func}) is larger than the IMU sampling period, we update the time offset and relaunch the system. In this case, the approximation and refinement processes can be skipped for saving computing resources.

\section{Experiments And Discussions}
\label{sec:experiments and discussions}
In this section, the performance of the proposed method is evaluated on synthetic sequences and public real-world datasets.
The results include the errors on the extrinsic spatial-temporal parameters, metric scales, gyroscope bias, accelerometer bias, and velocity.
For simplicity, all these errors are defined as scalars. The extrinsic translation and orientation errors are respectively the magnitude of the vectors, which indicate the difference between the ground-truth extrinsic parameters and the calibrated results. 
The ground-truth and the calibrated rotation matrices are represented in Euler angle vectors \emph{(yaw-pitch-roll} order). Similarly, the errors of time offset, gyroscope bias, accelerometer bias, and velocity are respectively the magnitude of the vectors showing the differences between the ground-truth values and the estimates.
The structure of this section is: Section \ref{subsec:Implement Detials} give the implement details of the proposed method. Section \ref{subsec:Simulation Experiments} reports the simulation results which evaluate the accuracy of calibrated extrinsic spatial-temporal parameters in the presence of various gyroscope and accelerometer noise. 
Section \ref{subsec:Real-World Experiments} analyzes the time offset influence, parameter convergence, and the overall VIVO acccuracy on real-world sequences.
All the experiments are carried out with an Intel CPU i7-4720HQ (8 cores @2.60GHz) laptop computer with 8GB RAM.

\subsection{Implement Details}
\label{subsec:Implement Detials}
Our method is implemented based on the monocular visual SLAM framework, termed ORB-SLAM \cite{mur2015orb, mur2017orb2}.
In particular, the \emph{Tracking} and \emph{Local Mapping} threads of this framework are adopted to respectively track the frame pose and deal with keyframes, which serve as monocular visual odometry front-end to collect keyframes.
The minimal number of required keyframes collected through the front-end is set to ten.
Once the calibrated extrinsic spatial parameters converge to stable values, the metric scales of keyframe poses and map points are immediately recovered, and the keyframe velocities are estimated.
The convergence criteria and velocity estimation method are similar to \cite{huang2020_tro_online_for_stereoVIO}.
At this point, the online initialization task can be considered to have completed.
As an option, a global bundle adjustment can be performed to further optimize all system states.
After this, we re-implement the \emph{Tracking} and \emph{Local Mapping} threads based on the proposed visual-inertial nonlinear optimization algorithm as described in Section \ref{sec:Visual-Inertial_State_Estimation}.
Note that we focus on the odometry technology, the \emph{Loop Closure} thread of ORB-SLAM is disabled. Besides, all the compared methods are run without loop closure.

\begin{figure*}[htbp]
	\centering \includegraphics[width=1.0\textwidth]{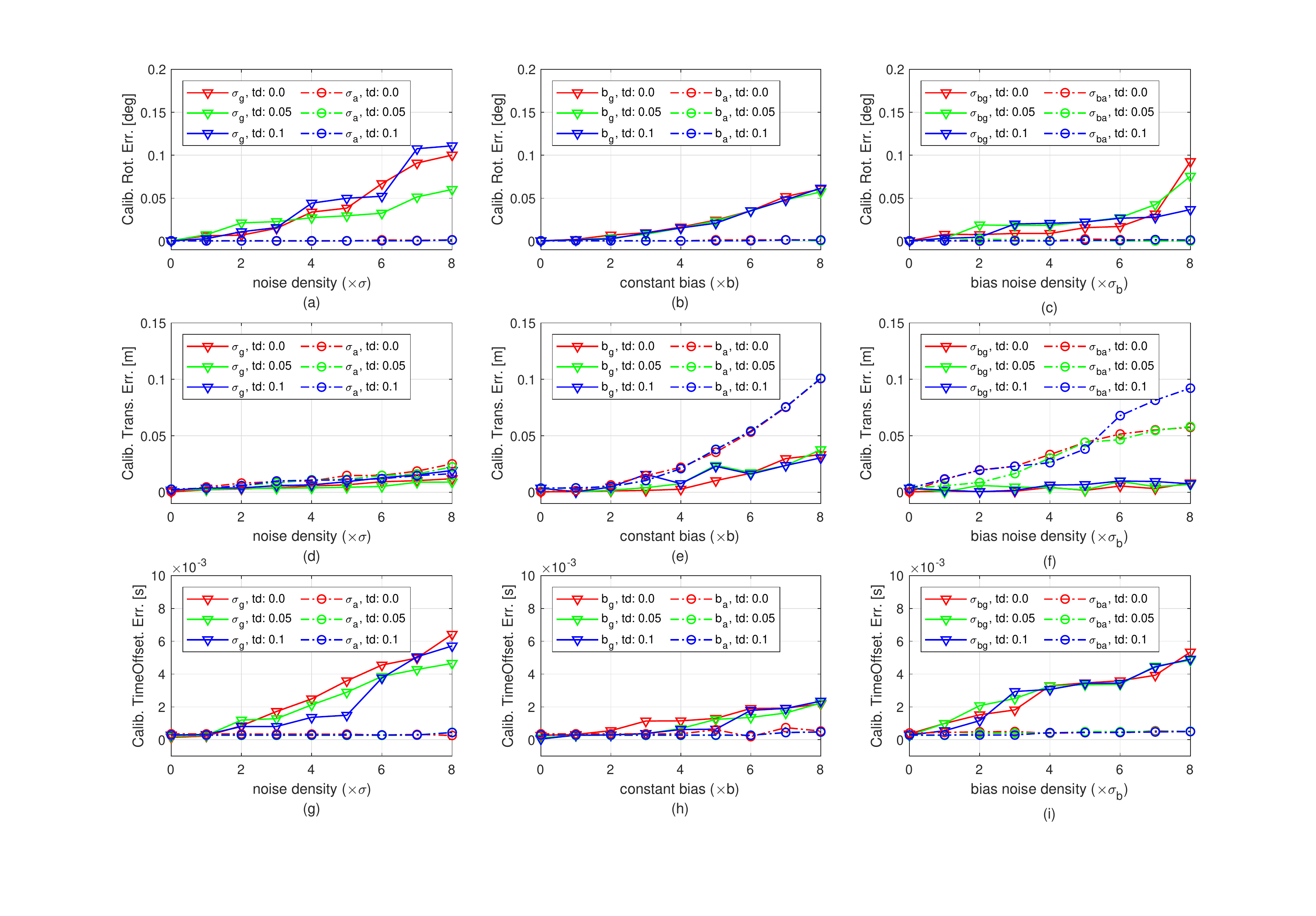}
	\vspace{-2.em}
	\caption{Extrinsic spatial-temporal calibration errors in the presence of various gyroscope and accelerometer noises.
		The vertical axes of (a)-(c), (d)-(f), and (g)-(i) subplots are, respectively, the calibration error of extrinsic camera-IMU rotation, translation, and time offset.
		The horizontal axes of the first column, the second column, and the third column are, respectively, the intensities of measurement noise density, constant bias, and bias ``diffusion'' random walk noise density. For instance, the label ``4'' in (a) means the gyroscope noise densities are set to $4\sigma_{g}$, while the other IMU noise parameters are set to zero. \textbf{Best viewed in color.}
	}
	\label{fig:Simulation_Calibrated_Rbc_Pbc_Td}
	\vspace{-0.1em}
\end{figure*}

\subsection{Simulation Experiments}
\label{subsec:Simulation Experiments}
In this experiment, a synthetic sequence that an IMU following a circular trajectory of a 3 m radius with a sinusoidal vertical motion is designed. The total length of the trajectory is 25.527 m.
The simulated sampling rates of IMU and camera are 200 Hz and 20 Hz respectively.
The IMU outputs are generated by computing the analytical derivatives of the parametric trajectory and adding white noises and slow time-varying biases.\footnote{The simulated IMU parameters are set as: Sampling rate: 200 Hz.
	 Gyroscope and accelerometer continuous-time noise densities: $\sigma_{g}=0.00017 \ \mathrm{rad/(s\sqrt{Hz})}$, $\sigma_{a}=0.002 \ \mathrm{m/(s^2\sqrt{Hz})}$. Constant biases: $\mathbf{b}_{g}=[-0.0023,0.0249, 0.0817] \ \mathrm{rad/s}$, $\mathbf{b}_{a}=[-0.0236,0.1210,0.0748] \ \mathrm{m/s^2}$.
	Bias ``diffusion" random walk noise densities: $\sigma_{bg}=0.00002 \
	\mathrm{rad/(s^2\sqrt{Hz})}$, $\sigma_{ba}=0.003 \ \mathrm{m/(s^3\sqrt{Hz})}$.
	We highlight that these basic parameters are similar to the ground-truth values provided by EuRoC dataset\cite{burri2016euroc}, therefore they are meaningful in practical application.
}
The camera\footnote{The simulated camera intrinsic parameters are set as: Sampling rate: 20 Hz. $(\mathrm{width}, \mathrm{height}) = (640, 640)$, $(f_x, f_y) = (460, 460)$, $(c_x, c_y) = (255, 255)$. The maximum number of feature points observed by one image is limited to 500 for saving computing resources.}
poses are obtained by transforming using customized camera-IMU extrinsic spatial parameters.\footnote{The simulated camera-IMU extrinsic parameters are set as: $\mathbf{R}^b_c = [180.0, 0.0, 0.0]$ deg for rotation and $\mathbf{p}^b_c = [0.1, 0.04, 0.03]$ m for translation. 
}
To simulate the effect of sensor asynchronous, we manually add time offsets to the camera timestamps. Therefore, the IMU and camera measurements are misaligned.
In the experiment, we test the performance on different time offsets, i.e., 0 ms, 50 ms, and 100 ms.
For each parameter setting, we generate 25 sequences and plot the median result.

The camera-IMU extrinsic spatial-temporal calibration errors in the presence of various IMU noises are shown in Fig. \ref{fig:Simulation_Calibrated_Rbc_Pbc_Td}. The horizontal axes of the subplots in the figure represent the noise intensities. The vertical axes are the calibration errors.
As shown in Fig. \ref{fig:Simulation_Calibrated_Rbc_Pbc_Td}(a)-(c) and (g)-(i), the calibration errors of extrinsic rotation and time offset grow with the increase of measurement noise, bias, and bias random walk noise of gyroscope.
However, we find that the calibration results of extrinsic rotation are satisfying since the maximum errors are smaller than $0.15^\circ$ even in the largest gyroscope noises.
The calibrated time offsets are also satisfying, i.e., smaller than the IMU sampling period (5 ms), when the gyroscope noise density and bias noise density are smaller than $7 \sigma_{g}$ and $7 \sigma_{bg}$.
The results also show that the fluctuations of the extrinsic rotation and time offset errors against the accelerometer noises are ignorable. This is reasonable since only the gyroscope-related parameters are involved in the first step of the three-step initialization process (see (\ref{e_Rbc_td_gyroBias})).
The accelerometer noises have little impact on extrinsic rotation and time offset calibration when performing global/local optimizations. The impact can be ignored since the optimizations estimate small correction.


Fig. \ref{fig:Simulation_Calibrated_Rbc_Pbc_Td}(d)-(f) show the calibration extrinsic translation errors against with gyroscope and accelerometer noises.
As shown in Fig. \ref{fig:Simulation_Calibrated_Rbc_Pbc_Td}(d), the calibration error grows slowly with the increase of measurement noise density. The maximum error is 0.02511 m when the accelerometer bias is $8 \sigma_{a}$, which indicates that the extrinsic translation calibration is robust to IMU noise density.
The curves in Fig. \ref{fig:Simulation_Calibrated_Rbc_Pbc_Td}(e) show that a large accelerometer bias will lead to poor calibration results.
In Fig. \ref{fig:Simulation_Calibrated_Rbc_Pbc_Td}(f), the curves show that the extrinsic translation calibration is robust to gyroscope bias noise density.
Although the calibration error grows with the increase of constant biases and accelerometer bias noise density, the results are still satisfying, i.e., 0.025 m errors, when the constant biases are smaller than $6 \mathrm{b}_{g}$ and $4 \mathrm{b}_{a}$, and the accelerometer bias noise density is smaller than $3 \sigma_{ba}$.

Note that in the simulation experiment, the calibration performance is evaluated on different time offsets.
We can find that the results plotted in different colors are not much different.
This phenomenon shows that the extrinsic parameter calibration capability of the proposed method is robust to different temporal misalignment.

\begin{figure}[htbp]
	\hspace{-0.7em}
	\includegraphics[width=0.5\textwidth]{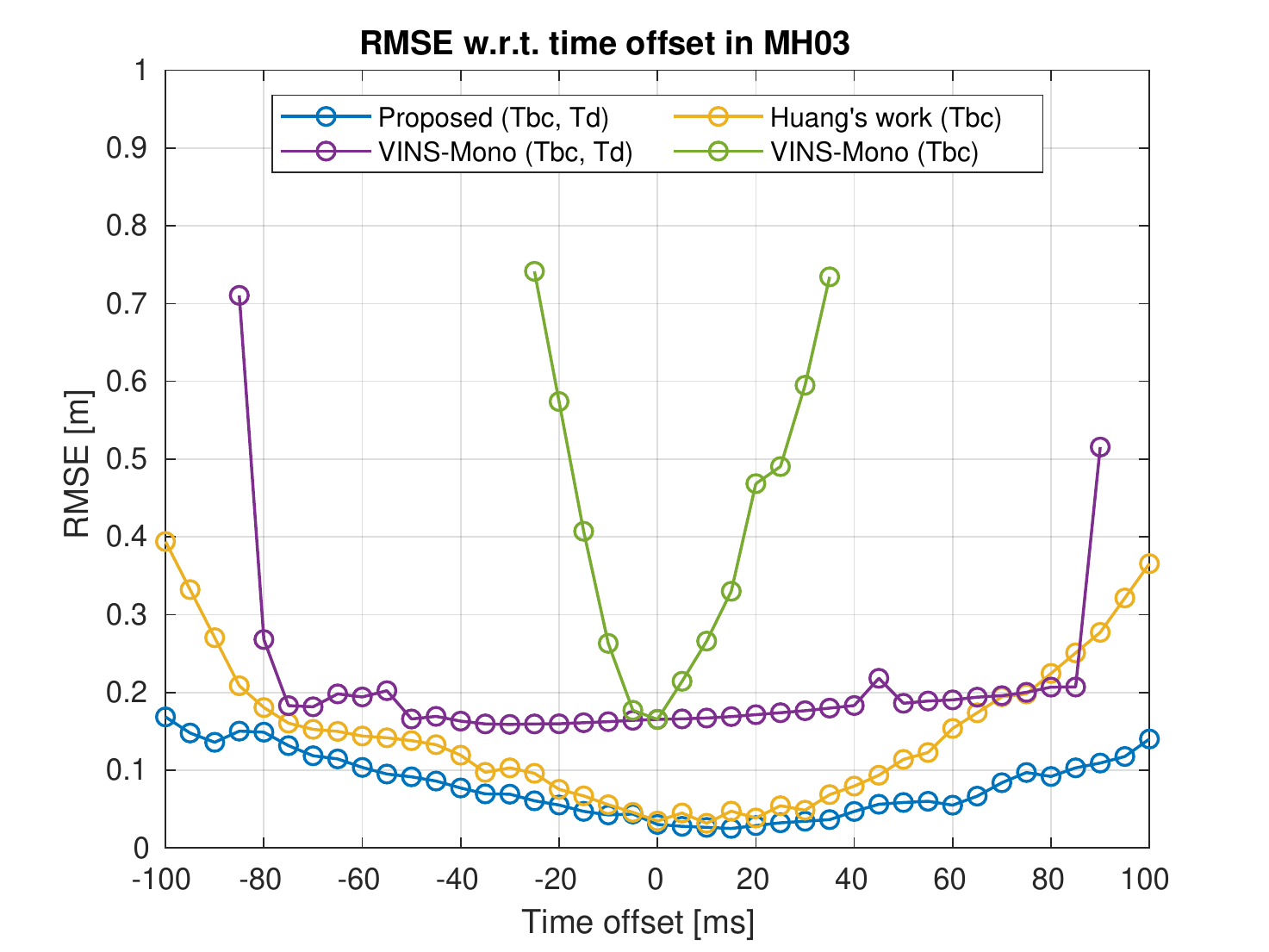}
	\vspace{-1.5em}
	\caption{Comparison of the trajectory accuracy with respect to different time offsets in MH\_03 sequence. The x-axis shows the predefined time offset, and the y-axis shows the absolute root mean square error (RMSE).
		The lines shown in figure are: ``Proposed (Tbc, Td)'' -- results of the proposed method, which has capability of spatial-temporal calibration;
		``Huang's work (Tbc)'' -- results of our earlier work \cite{huang2018online}, which has capability of spatial calibration; ``VINSMono (Tbc, Td)'' -- results of VINSMono \cite{qin2018vins} that was launched under ``no extrinsic parameters'' and ``estimate time offset'' configurations; ``VINSMono (Tbc)'' -- results of VINSMono that was launched under ``no extrinsic parameters'' configuration. \textbf{Best viewed in color.}
	}
	\label{fig:RMSE_wrt_TimeOffset_In_MH03}
\end{figure}

\subsection{Real-World Experiments}
\label{subsec:Real-World Experiments}

\subsubsection{Dataset}
\label{subsubsec:Dataset}
The real-world performance of the proposed method is evaluated on the EuRoC dataset \cite{burri2016euroc}.
By considering the illumination, texture, fast/slow motions or motion blur, the sequences can be classified into \emph{easy}, \emph{medium}, and \emph{difficult} sets.
It not only provides accurate ground-truth of flying-trajectories, velocities of IMU body, gyroscope bias, and accelerometer bias, but also offers accurate camera-IMU extrinsic spatial parameters\footnote{
	The ground-truth extrinsic parameters provided by the dataset were calibrated by the \emph{Kalibr}\cite{rehder2016extending,furgale2013unified,furgale2012continuous,maye2013self}  toolbox,
	with [89.147953, 1.476930, 0.215286] degree in yaw, pitch, roll directions for $\mathbf{R}^b_c$, and [-0.021640, -0.064677, 0.009811] meter in x, y, z directions for $\mathbf{p}^b_c$.}.
It is also well known that the images and IMU measurements are strictly hardware time-synchronized and logged at 20 and 200 Hz.
These characteristics make the dataset become an ideal choice for evaluating the accuracy of extrinsic parameters calibration and initial values estimation.
In the following experiments, we manually add a fixed millisecond value to image timestamps to conduct time-shifted sequences, such that there is a fixed time offset between IMU and camera measurements.
The time-shifted sequences are used to test the proposed algorithm and other methods.

\subsubsection{Time Offset Influence}
\label{subsubsec:Time Offset Influence}
In this experiment, the influence of time offset on visual-inertial odometry is studied.
As shown in Fig. \ref{fig:RMSE_wrt_TimeOffset_In_MH03}, we added the time offsets from -100 to 100 ms on MH\_03 sequence, and tested the time-shifted sequences with Huang's work \cite{huang2018online}, VINS-Mono \cite{qin2018vins}, and the proposed method, respectively.
Huang's work is our earlier work that can online calibrate extrinsic spatial parameters, whereas it cannot calibrate the extrinsic temporal parameter.
The results of this work are shown in yellow line with a legend of ``Huang's work (Tbc)''.
VINS-Mono is a state-of-the-art monocular VIO algorithm with an online spatial-temporal calibration ability. It provides three configurations, i.e., ``\textit{with extrinsic parameters}'', ``\textit{have initial extrinsic guess}'', and ``\textit{no extrinsic parameters}'', about spatial calibration and, one configuration, i.e., ``\textit{estimate time offset}'' about temporal calibration.
For the ``\textit{no extrinsic parameters}'' configuration, the public project integrates the authors' studies on automatic estimator initialization and online extrinsic spatial calibration, which can be found in \cite{qin2017robust} and \cite{yang2017monocular}.
For the ``\textit{estimate time offset}'' configuration, it integrates the authors' study on online temporal calibration, which can be found in \cite{qin2018online}.
Here, we launched VINS-Mono under ``\textit{no extrinsic parameters}'' and ``\textit{estimate time offset}'' configurations for fairly comparing with the proposed method. The results are colored in purple with a legend of ``VINS-Mono (Tbc, Td)''.
For a comprehensive comparison, VINS-Mono was also launched merely under the ``\textit{no extrinsic parameters}'' configuration, while the temporal calibration ability was disabled.
The results of this configuration are colored in green with a legend of ``VINS-Mono (Tbc)''.
%
Note that all these methods were launched without given any initial guess about extrinsic spatial or temporal parameters. All results are the median over 25 tests.
In this experiment, we find out that the results of ``VINS-Mono (Tbc, Td)'' are of huge errors when the time offset surpasses 90 or -85 ms, hence these results are not plotted for limiting the range of y-axis. Similarly, the results of ``VINS-Mono (Tbc)'' are not plotted when time offset surpasses 35 or -25 ms due to huge errors.

\begin{figure}[tbp]
	\includegraphics[width=0.5\textwidth]{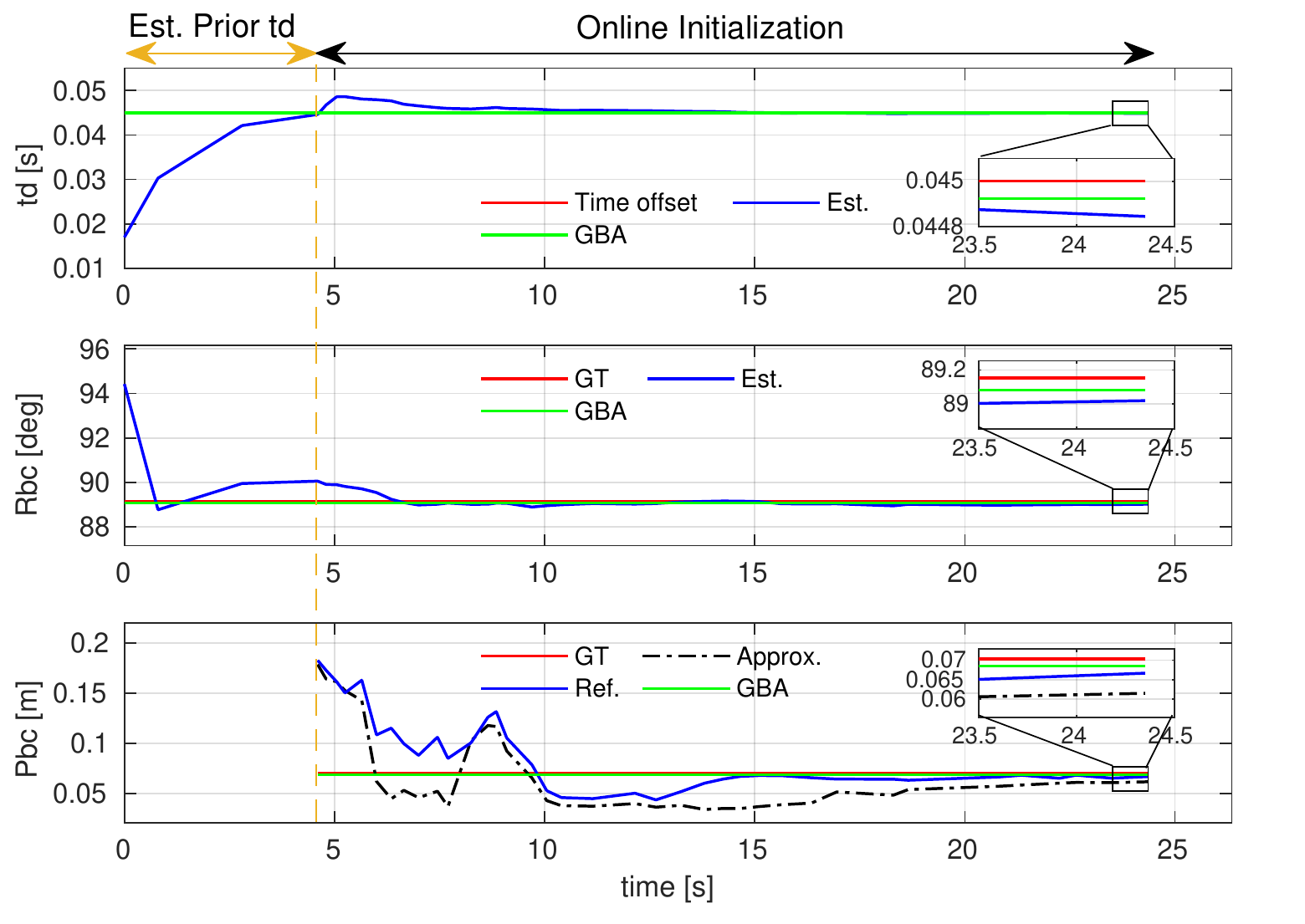}
	\vspace{-1.8em}
	\caption{Calibrated extrinsic spatial and temporal parameters in the V2\_01 sequence. The term ``Est. Prior td'' denotes the stage in which we update the time offset and relaunch the system due to large time increment estimation (see Section \ref{subsec:Compensation of Time Offset and Initialization Trick} for details).
		Abbreviations: Est. -- estimated; GBA -- global bundle adjustment; GT -- ground-truth; Approx. -- approximation; Ref. -- refinement. \textbf{Best viewed in color.}
	}
	\label{fig:TimeDelay_Rbc_Pbc}
\end{figure}

As shown in Fig. \ref{fig:RMSE_wrt_TimeOffset_In_MH03}, the accuracy of the proposed method is better than Huang's work, especially when the time offset surpasses 30 or -20 ms. The performance of ``VINS-Mono (Tbc, Td)'' is also much better than ``VINS-Mono (Tbc)''.
These phenomena demonstrate that the temporal calibration significantly benefits overall performance and thus it is necessary to perform a temporal calibration.
Comparing the blue with purple lines, it is obvious that the proposed method achieves much lower absolute trajectory RMSE than ``VINS-Mono (Tbc, Td)'' in all predefined time offsets.
All the trajectory errors of the proposed method are below 0.17 meters, which proves our approach is able to estimate accurate enough sensor pose under a wide range of temporal offsets.
On the contrary, ``VINS-Mono (Tbc, Td)'' achieves fairly consistent accuracy when time offset is within -75 to 85 ms, whereas the performance deteriorates dramatically when the time offset increases.
This might be because the authors used feature velocity for modeling and compensating the temporal misalignment, by assuming that an image feature moves at an approximately constant velocity on the image plane in a short period. However, when the time offset is large, e.g., larger than 85 ms, this assumption might be violated. Thus, the feature velocity can not well compensate for the temporal misalignment, which leads to poor performance.


\begin{figure}[t]
\includegraphics[width=0.5\textwidth]{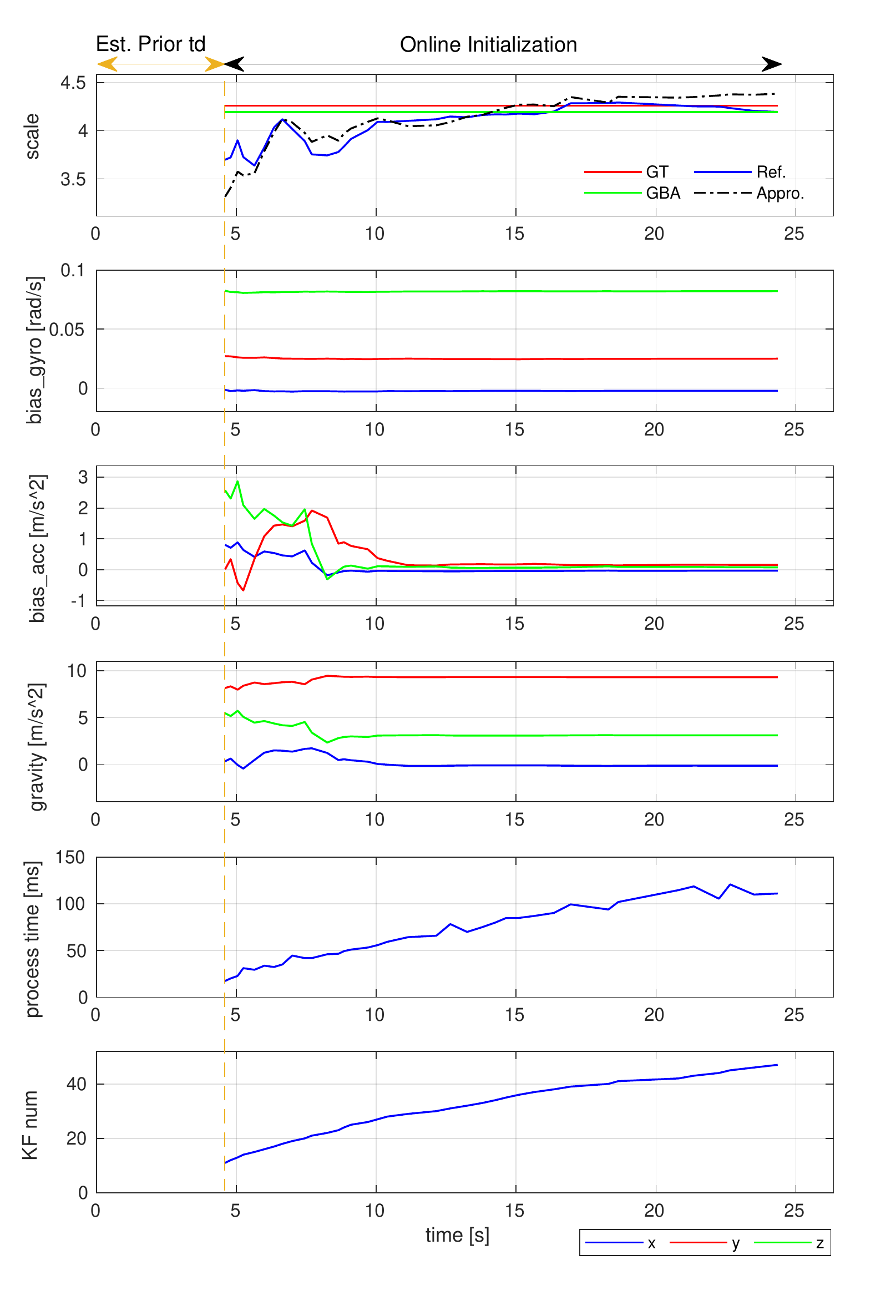}
	\vspace{-3.em}
	\caption{Estimated scale, gyroscope bias (bias\_gyro), accelerometer bias (bias\_acc), gravity, processing time expended by each execution, and keyframe number (KF num) in the V2\_01 sequence.
	}
	\vspace{-.5em}
	\label{fig:Scale_Biases_Gravity_Processtime_KFnum}
\end{figure}

\subsubsection{Convergence Performance}
\label{subsubsec:Convergence Performance}
In this experiment, the convergence performance of extrinsic spatial-temporal calibration and initial value estimation is analyzed on V2\_01 sequence, with a predefined time offset of 45 ms.
The time varied characteristic curves of the calibrated spatial-temporal results are shown in Fig. \ref{fig:TimeDelay_Rbc_Pbc}, and the curves of estimated initial values are shown in Fig. \ref{fig:Scale_Biases_Gravity_Processtime_KFnum}.
Note that at the beginning, as described in Section \ref{subsec:Compensation of Time Offset and Initialization Trick}, if a new time increment estimated by (\ref{gyroBias_Rbc_td_min_func}) is larger than the IMU sampling period, our system will be relaunched after the time offset is updated. This stage is termed as ``Est. Prior td''. Since the approximation and refinement processes are skipped in this stage, the corresponding curves of extrinsic translation and initial values are not plotted in Fig. \ref{fig:TimeDelay_Rbc_Pbc} and Fig. \ref{fig:Scale_Biases_Gravity_Processtime_KFnum}.

The curves of time offset, extrinsic rotation, and gyroscope bias show that these parameters can quickly converge to stable values within around 7 seconds.
And, the stable values are very close to ground-truth. These phenomena demonstrate that the first step of the introduced three-step process described in Section \ref{subsec:Gyroscope Bias Estimation, Extrinsic Rotation and Temporal Calibration} can effectively calibrate the extrinsic rotation and time offset.
As shown in the curves of extrinsic translation and scale, we can find that the refined results (i.e., blue lines) are better than the approximated ones (i.e., black dash-dot lines). The results are even better after global optimization (i.e., green lines).
This phenomenon indicates that the coarse-to-fine strategy introduced in  Section \ref{subsec:Scale_Gravity and Translation Approximation} and \ref{subsec:Accelerometer Bias Estimation and Scale Gravity and Translation Refinement} exhibits good performance.
Besides, the proposed nonlinear optimization algorithm can further improve the system states.

It is worth noting that the curves of accelerometer bias and gravity suffer severe oscillation in the first few seconds. This is because the platform did not have enough excitation on at least two independent axes of the sensor suite at the beginning, which made the accelerometer bias and the gravity indistinguishable.
With new keyframes coming, they could be well estimated.
The keyframe number and the processing time expended by each execution are also plotted in Fig. \ref{fig:Scale_Biases_Gravity_Processtime_KFnum}.
It shows that the processing time is approximately linear to the number of keyframes, indicating that our method has linear time complexity.

\begin{figure}[tbp]
	\hspace{-0.7em}
	\includegraphics[width=0.5\textwidth]{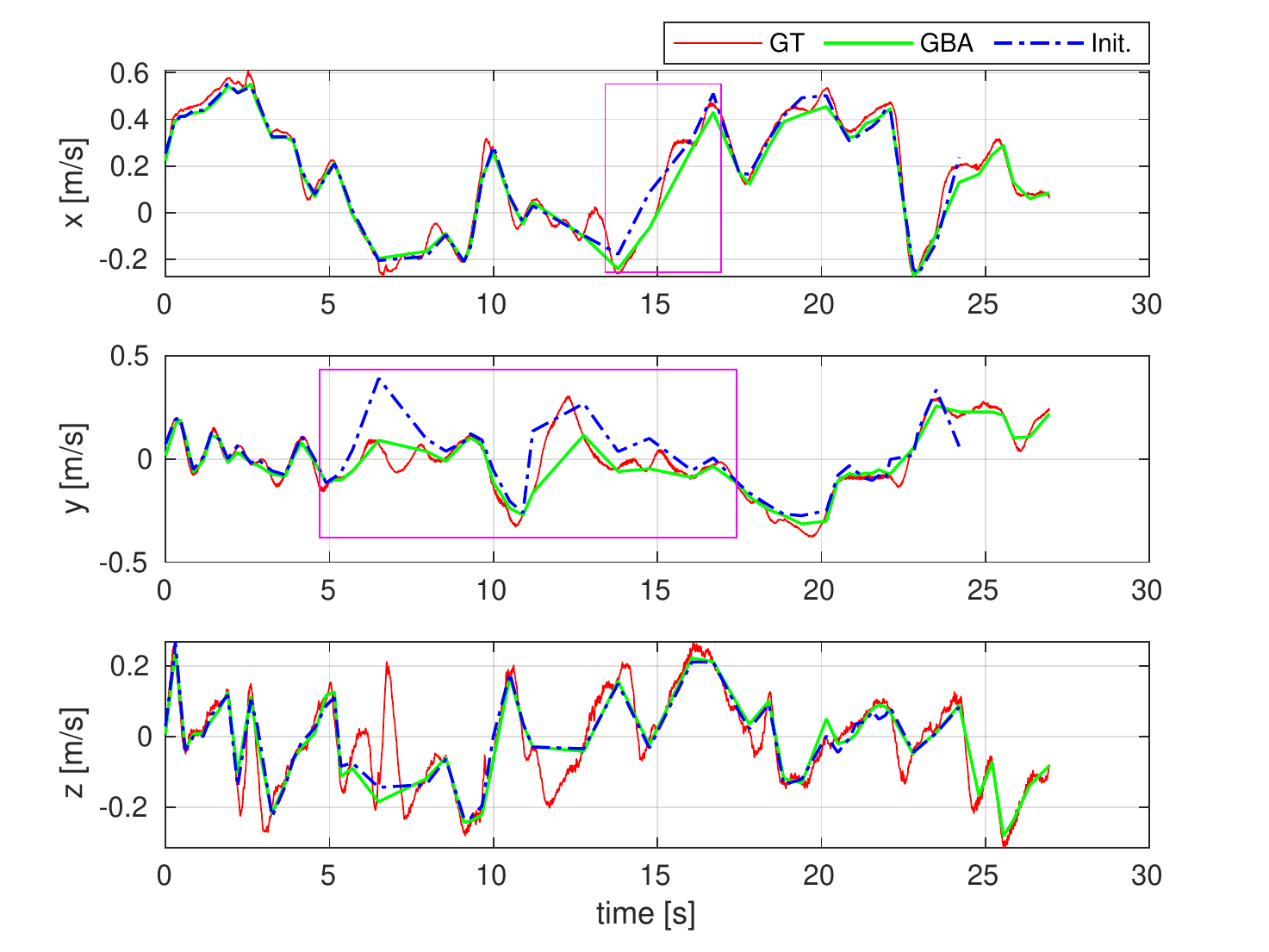}
	\vspace{-1.5em}
	\caption{Estimated IMU body velocities in x, y, z directions in the V2\_01 sequence. Red line: ground-truth velocity (GT); blue dash-dot line: the velocity estimated by online initialization (Init.); green line: the velocity optimized by global BA (GBA). \textbf{Best viewed in color.}
	}
	\label{fig:velocity}
\end{figure}

\subsubsection{Velocity Estimation}
\label{subsubsec:Velocity Estimation}
The curves of estimated IMU body velocity are plotted in Fig. \ref{fig:velocity}.
Since the estimates and ground-truth are expressed in different coordinate systems, the estimates are rotated to best fit with the ground-truth.
It can be seen that the initialization results (i.e., blue dash-dot lines) are consistent with the ground-truth (i.e., red lines) to some extend. This indicates that the velocities can be on the whole well estimated by the proposed method.
As shown in the pink rectangle regions, we find that the accuracy of velocity estimation can be further improved by performing a nonlinear global optimization.
The root mean square error of the initialized and optimized velocities are respectively 0.093 m/s and 0.046 m/s.
Fig. \ref{fig:velocity} also shows that the scale can be correctly estimated since otherwise, the magnitude of estimated velocity would differ from the ground-truth.

\begin{table*}[tbp]
  \centering
  \begin{threeparttable}
  \caption{Spatial-Temporal Calibration Errors and Keyframe Trajectory Accuracy in EuRoC Dataset.\tnote{1}}
  \tabcolsep0.7em
    \begin{tabular}{cccccccccccccccc}
    \toprule
      &   & \multicolumn{4}{c}{\textbf{VINS-Mono}\cite{qin2018vins}} &   & \multicolumn{4}{c}{\textbf{Feng \textit{et. al}}\cite{feng2019online}} &   & \multicolumn{4}{c}{\textbf{Ours}} \\
    \cmidrule{3-6}\cmidrule{8-11}\cmidrule{13-16}   & \textbf{time offset}  & \textbf{e\_Rbc}  & \textbf{e\_Pbc}  & \textbf{e\_td} & \textbf{RMSE} &  & \textbf{e\_Rbc}  & \textbf{e\_Pbc}  & \textbf{e\_td} & \textbf{RMSE} & & \textbf{e\_Rbc}  & \textbf{e\_Pbc}  & \textbf{e\_td} & \textbf{RMSE}\\
     & (ms) & (deg) & (m) & ms & (m) & & (deg) & (m) & ms & (m) & & (deg) & (m) & ms & (m)\\
    \midrule
    \multicolumn{1}{c}{\multirow{3}[2]{*}{V1\_01}} & 0 & 0.566 & 0.020 & -1.52 & 0.096 &   & 0.583 & 0.022 & -0.15 & 0.073 &   & \textbf{0.392} & \textbf{0.008} & \textbf{0.00} & \textbf{0.047} \\
      & 50 & 0.571 & 0.016 & -1.77 & 0.084 &   & 0.588 & 0.023 & \textbf{-0.21} & 0.073 &   & \textbf{0.136} & \textbf{0.008} & 0.84 & \textbf{0.045} \\
      & 100 & 0.624 & \textbf{0.010} & -3.23 & \textbf{0.067} &   & 0.577 & 0.022 & \textbf{-0.15} & 0.077 &   & \textbf{0.321} & 0.037 & 0.24 & 0.074 \\
\cmidrule{1-6}\cmidrule{8-11}\cmidrule{13-16}    \multicolumn{1}{c}{\multirow{3}[2]{*}{V1\_02}} & 0 & 0.534 & 0.046 & -0.57 & 0.091 &   & 0.563 & \textbf{0.019} & -0.09 & 0.118 &   & \textbf{0.128} & 0.021 & \textbf{-0.03} & \textbf{0.025} \\
      & 50 & 0.623 & 0.018 & -0.88 & 0.070 &   & 0.559 & 0.019 & -0.10 & 0.116 &   & \textbf{0.209} & \textbf{0.003} & \textbf{0.08} & \textbf{0.062} \\
      & 100 & 0.672 & 0.018 & -1.53 & \textbf{0.064} &   & 0.569 & 0.021 & \textbf{-0.10} & 0.143 &   & \textbf{0.173} & \textbf{0.015} & 2.37 & 0.115 \\
\cmidrule{1-6}\cmidrule{8-11}\cmidrule{13-16}    \multicolumn{1}{c}{\multirow{3}[2]{*}{V1\_03}} & 0 & 0.515 & 0.017 & -0.35 & \multicolumn{1}{c}{--- \tnote{2}} &   & 0.507 & 0.013 & \textbf{-0.33} & 0.118 &   & \textbf{0.165} & \textbf{0.012} & -0.34 & \textbf{0.012} \\
      & 50 & 0.547 & \textbf{0.010} & -0.87 & 0.407 &   & 0.508 & 0.016 & \textbf{-0.33} & 0.121 &   & \textbf{0.276} & 0.045 & -3.19 & \textbf{0.066} \\
      & 100 & \multicolumn{1}{c}{---} & \multicolumn{1}{c}{---} & \multicolumn{1}{c}{---} & \multicolumn{1}{c}{---} &   & 0.513 & \textbf{0.014} & \textbf{-0.39} & \textbf{0.093} &   & \textbf{0.194} & 0.023 & -2.69 & 0.102 \\
\cmidrule{1-6}\cmidrule{8-11}\cmidrule{13-16}    \multicolumn{1}{c}{\multirow{3}[2]{*}{V2\_01}} & 0 & 0.471 & 0.024 & -1.11 & 0.065 &   & 0.491 & 0.023 & \textbf{-0.33} & 0.099 &   & \textbf{0.280} & \textbf{0.017} & -1.01 & \textbf{0.019} \\
      & 50 & 0.573 & 0.021 & -0.95 & 0.053 &   & 0.457 & 0.025 & \textbf{-0.29} & 0.088 &   & \textbf{0.360} & \textbf{0.019} & 0.30 & \textbf{0.021} \\
      & 100 & 0.645 & \textbf{0.019} & -2.32 & \textbf{0.034} &   & 0.513 & 0.022 & -0.36 & 0.082 &   & \textbf{0.489} & 0.037 & \textbf{0.28} & 0.047 \\
\cmidrule{1-6}\cmidrule{8-11}\cmidrule{13-16}    \multicolumn{1}{c}{\multirow{3}[2]{*}{V2\_02}} & 0 & 0.599 & \textbf{0.014} & -0.40 & 0.090 &   & 0.553 & 0.020 & \textbf{-0.09} & 0.099 &   & \textbf{0.065} & 0.020 & -1.16 & \textbf{0.028} \\
      & 50 & 0.651 & \textbf{0.013} & -0.49 & 0.144 &   & 0.558 & 0.020 & \textbf{-0.09} & 0.089 &   & \textbf{0.133} & 0.021 & 0.17 & \textbf{0.051} \\
      & 100 & 0.581 & \textbf{0.009} & -0.79 & \multicolumn{1}{c}{---} &   & 0.558 & 0.020 & \textbf{-0.09} & 0.100 &   & \textbf{0.254} & 0.018 & 1.75 & \textbf{0.097} \\
\cmidrule{1-6}\cmidrule{8-11}\cmidrule{13-16}    \multicolumn{1}{c}{\multirow{3}[2]{*}{MH\_01}} & 0 & 0.552 & \textbf{0.018} & -0.68 & 0.241 &   & 0.501 & 0.018 & \textbf{-0.16} & \textbf{0.080} &   & \textbf{0.403} & 0.021 & 1.84 & 0.122 \\
      & 50 & 0.556 & \textbf{0.014} & -0.85 & 0.247 &   & 0.505 & 0.015 & \textbf{-0.12} & 0.119 &   & \textbf{0.415} & 0.018 & -0.13 & \textbf{0.053} \\
      & 100 & 0.533 & 0.025 & -1.49 & 0.366 &   & 0.481 & \textbf{0.015} & -0.12 & 0.111 &   & \textbf{0.364} & 0.035 & \textbf{0.11} & \textbf{0.098} \\
\cmidrule{1-6}\cmidrule{8-11}\cmidrule{13-16}    \multicolumn{1}{c}{\multirow{3}[2]{*}{MH\_02}} & 0 & 0.537 & \textbf{0.010} & -0.93 & 0.292 &   & 0.621 & 0.014 & \textbf{-0.29} & 0.082 &   & \textbf{0.149} & 0.030 & 0.60 & \textbf{0.021} \\
      & 50 & 0.512 & \textbf{0.008} & -1.25 & 0.277 &   & 0.624 & 0.014 & -0.34 & \textbf{0.086} &   & \textbf{0.316} & 0.016 & \textbf{-0.23} & 0.097 \\
      & 100 & 0.556 & \textbf{0.014} & -1.05 & \multicolumn{1}{c}{---} &   & 0.634 & 0.015 & -0.21 & 0.074 &   & \textbf{0.316} & 0.032 & \textbf{-0.01} & \textbf{0.060} \\
\cmidrule{1-6}\cmidrule{8-11}\cmidrule{13-16}    \multicolumn{1}{c}{\multirow{3}[2]{*}{MH\_03}} & 0 & 0.619 & \textbf{0.019} & -0.82 & 0.192 &   & 0.619 & 0.022 & \textbf{-0.01} & 0.161 &   & \textbf{0.225} & 0.023 & -2.03 & \textbf{0.030} \\
      & 50 & 0.671 & \textbf{0.014} & -1.20 & 0.189 &   & 0.627 & 0.024 & \textbf{-0.05} & 0.133 &   & \textbf{0.211} & 0.028 & -2.36 & \textbf{0.058} \\
      & 100 & 1.132 & 0.035 & -2.77 & \multicolumn{1}{c}{---} &   & 0.607 & \textbf{0.020} & \textbf{-0.09} & 0.173 &   & \textbf{0.304} & 0.036 & -1.44 & \textbf{0.140} \\
\cmidrule{1-6}\cmidrule{8-11}\cmidrule{13-16}    \multicolumn{1}{c}{\multirow{3}[2]{*}{MH\_04}} & 0 & 0.560 & 0.022 & -1.15 & 0.372 &   & 0.554 & \textbf{0.019} & 0.11 & 0.197 &   & \textbf{0.044} & 0.019 & \textbf{0.00} & \textbf{0.157} \\
      & 50 & 0.558 & 0.013 & -1.46 & 0.487 &   & 0.521 & \textbf{0.013} & \textbf{0.17} & \textbf{0.178} &  & \textbf{0.349} & 0.029 & -0.31 & 0.240 \\
      & 100 & 0.468 & \textbf{0.007} & -3.12 & 0.331 &   & 0.512 & 0.018 & \textbf{-0.03} & \textbf{0.143} &  & \textbf{0.250} & 0.023 & 1.28 & 0.215 \\
\cmidrule{1-6}\cmidrule{8-11}\cmidrule{13-16}    \multicolumn{1}{c}{\multirow{3}[2]{*}{MH\_05}} & 0 & 0.538 & 0.020 & -1.26 & 0.309 &   & 0.605 & \textbf{0.013} & \textbf{-0.09} & \textbf{0.162} & & \textbf{0.107} & 0.016 & -0.47 & 0.204 \\
      & 50 & 0.547 & 0.017 & -1.49 & 0.299 &   & 0.509 & \textbf{0.010} & \textbf{-0.20} & \textbf{0.207} &  & \textbf{0.178} & 0.012 & 0.44 & 0.254 \\
      & 100 & 0.435 & 0.088 & -2.20 & 1.141 &   & 0.552 & \textbf{0.017} & \textbf{-0.17} & \textbf{0.205} &  & \textbf{0.357} & 0.026 & -0.62 & 0.214 \\
    \bottomrule
    \end{tabular}%
  \begin{tablenotes}
        \footnotesize
        \item[1] All the results of our method are the median over 25 tests in each sequence of EuRoC dataset.
        \item[2] ``---'' means that the tracking is lost at some point and a significant portion of the sequence is not processed by the system.
    \end{tablenotes}
  \label{tab:whole_EuRoC_dataset_performance}%
  \end{threeparttable}
\end{table*}%

\subsubsection{Accuracy on the Whole Dataset}
\label{subsubsec:Accuracy on the Whole Dataset}

In this experiment, we compared the proposed method with VINS-Mono \cite{qin2018vins} and Feng's work \cite{feng2019online}. VINS-Mono was launched under ``\textit{no extrinsic parameters}'' and ``\textit{estimate time offset}'' configurations.
Feng's work introduced an online spatial-temporal calibration method for monocular direct VIO.
The time offset was set to 0 ms, 50 ms, and 100 ms for comparison.
The errors of the calibrated spatial-temporal parameters and the absolute translational RMSE of keyframe trajectories are shown in Table \ref{tab:whole_EuRoC_dataset_performance}, in which the results of VINS-Mono and Feng's work were cloned from \cite{feng2019online}.
It can be seen that the average error of extrinsic rotation calibrated by our method is about 0.252 degrees, which performs much more accurately than VINS-Mono (0.584 degrees) and Feng's work (0.552 degrees).
The average absolute errors of extrinsic translation and time offset of our method are respectively 0.022 m and 0.877 ms, which are competitive compared with the other two methods.
In addition, the results show that the trajectory estimated by our method has the highest accuracy on most sequences.

\section{Conclusions}
\label{sec:Conclusions}
This work studied the online initialization and self-calibration problem for bootstrapping the monocular visual-inertial odometry.
By introducing the short-term motion interpolation algorithms for camera and IMU, we found out that the temporal misalignment problem could be well solved.
In particular, the extrinsic spatial-temporal parameters between camera and IMU, and the initial metric scale, velocity, gravity, and IMU biases could be simultaneously estimated using a three-step process.
Besides, by considering the time offset in the nonlinear optimization, all the system states could be further optimized.
Since the proposed method does not rely on any prior knowledge about the mechanical or temporal configuration, it is suitable for the VIO sensors where the extrinsic spatial parameters were unknown or the timestamps were not well synchronized.
The performance of our method is evaluated on both the synthetic sequences and the public dataset. The results show that the initial values and extrinsic parameters can be accurately estimated and converge in a short time.
The trajectory of the platform can also be estimated by the introduced nonlinear optimization, and it exhibits a competitive accuracy compared with the popular VINS-Mono method.

\appendix
\subsection{Preliminaries}
\label{appendix: preliminaries}
In this section, we give some background geometric concepts that will be used in the following sections.

\subsubsection{First-order approximation}
The exponential map of a rotational vector $\phi \in \mathfrak{so}(3)$  is equivalent to a standard matrix exponential (Rodrigues' rotation formula):
\begin{equation}\label{exponential_map}
\mathrm{exp}(\phi^\wedge) = \mathbf{I}+\frac{\mathrm{sin}(\|\phi\|)}{\|\phi\|} \phi^\wedge + \frac{1-\mathrm{cos}(\|\phi\|)}{\|\phi\|^2}(\phi^\wedge)^2.
\end{equation}
A first-order approximation of Taylor expansion for the exponential map is:
\begin{equation}\label{exponential_map_first_order_appro}
\mathrm{exp}(\phi^\wedge) \approx \mathbf{I}  + \phi^\wedge.
\end{equation}
Note that one could also use $\mathrm{Pad\acute{e}}$ approximants \cite{baker1975essentials} for better approximating the exponential map.

\subsubsection{Adjoint property}
Give a Lie element $\tilde{\phi} = \mathrm{Log}(\tilde{\mathbf{R}})$ and a rotation $\mathbf{R}$, the adjoint property is:
\begin{equation}\label{adjoint_property}
  \mathbf{R}\tilde{\mathbf{R}}\mathbf{R}^T = \mathrm{exp} ( \mathbf{R} \tilde{\phi}^\wedge \mathbf{R}^T) = \mathrm{exp}((\mathbf{R} \tilde{\phi})^\wedge) = \mathrm{Exp}((\mathbf{R} \tilde{\phi}).
\end{equation}

\subsubsection{BCH linear approximation}
The BCH (Baker-Campbell-Hausdorff \cite{gilmore1974baker}) linear approximation for $\phi_1$ and $\phi_2$ in the Lie algebra of a Lie group is:
\begin{equation}\label{BCH_linear_approximation}
  \mathrm{Log}(\mathrm{Exp}(\phi_1) \mathrm{Exp}(\phi_2)) \approx \left\{ \!\!  \begin{array}{c}
                                                                 \mathbf{J}_l^{-1}(\phi_2) \phi_1 + \phi_2, \ \! \mathrm{if} \ \! \phi_1 \ \! \mathrm{is \ \! small}, \\ [2mm]
                                                                 \mathbf{J}_r^{-1}(\phi_1) \phi_2 + \phi_1, \ \! \mathrm{if} \ \! \phi_2 \ \! \mathrm{is \ \! small},
                                                               \end{array} \right.
\end{equation}
where $\mathbf{J}_l^{-1}(\cdot)$ and $\mathbf{J}_r^{-1}(\cdot)$ are the inverses of left-jacobian matrix $\mathbf{J}_l(\cdot)$ and right-jacobian matrix $\mathbf{J}_r(\cdot)$ respectively.
There we can have the additive operation of a small perturbation $\delta \phi$ on Lie algebra, as follows:
\begin{equation}\label{additive_operation_of_lie_algebra}
\begin{aligned}
  \mathrm{Exp}(\phi + \delta \phi) &\approx \mathrm{Exp}(\mathbf{J}_l(\phi) \delta \phi) \mathrm{Exp}(\phi) \\
   &\approx \mathrm{Exp}(\phi) \mathrm{Exp}(\mathbf{J}_r(\phi) \delta \phi)
\end{aligned}
\end{equation}

\subsection{State Update}
\label{appendix: State Update}
In this section, we provide the retraction expressions for updating the system states, as follows:
\begin{equation}\label{retractions}
  \begin{array}{cc}
    \mathbf{R}^w_{b_i} \leftarrow \mathbf{R}^w_{b_i} \mathrm{Exp}(\delta \phi_{b_i}), & \mathbf{R}^b_c \leftarrow \mathbf{R}^b_c \mathrm{Exp}(\delta \phi^b_c), \\[2mm]
    \mathbf{p}^w_{b_i} \leftarrow \mathbf{p}^w_{b_i} + \mathbf{R}^w_{b_i} \delta \mathbf{p}_{b_i} &  \mathbf{p}^b_c \leftarrow \mathbf{p}^b_c + \mathbf{R}^b_c \delta \mathbf{p}^b_c, \\[2mm]
    \mathbf{v}^w_{b_i} \leftarrow \mathbf{v}^w_{b_i} + \delta \mathbf{v}_{b_i}, \ \ \ \ \   &  \mathbf{p}^w_{k} \!\! \leftarrow \mathbf{p}^w_{k} + \delta \mathbf{p}^w_{k} , \ \ \  \\[2mm]
    \delta \mathbf{b}_{g_i} \leftarrow \delta \mathbf{b}_{g_i} + \tilde{\delta} \mathbf{b}_{g_i},  \ \ \ \ \  & t_d \leftarrow t_d + \delta t_d, \ \ \ \ \  \\[2mm]
    \delta \mathbf{b}_{a_i} \leftarrow \delta \mathbf{b}_{a_i} + \tilde{\delta} \mathbf{b}_{a_i}.  \ \ \ \ \  & \
  \end{array}
\end{equation}
Note that the retraction makes the residual errors in (\ref{nonlinear_optimization}) a function defined on a vector space, on which it is easy to compute Jacobians. Therefore, in the following sections, we derive the Jacobians w.r.t. the vectors $\delta \phi_{b_i}$, $\delta \mathbf{p}_{b_i}$, $\delta \mathbf{v}_{b_i}$, $\tilde{\delta} \mathbf{b}_{g_i}$, $\tilde{\delta} \mathbf{b}_{a_i}$, $\delta \phi^b_c$, $\delta \mathbf{p}^b_c$, $\delta \mathbf{p}^w_{k}$, and $\delta t_d$.

\subsection{Jacobians of Feature Residual Errors}
\label{appendix: Jacobians of Reprojection Errors}
We define the residual error between the reprojection of $\mathbf{p}^w_k$ and the pixel location $\mathbf{u}^i_k$ of matched feature as $\mathbf{r}_{\mathcal{C}_{ik}} = \mathbf{u}^i_k - \pi(\mathbf{p}^{c_i}_k)$, in which $\mathbf{p}^{c_i}_k$ is the $k$th map point that observed by the $i$th keyframe and expressed in the camera frame.
Since $\mathbf{b}_{a_i}$ does not appear in $\mathbf{r}_{\mathcal{C}_{ik}}$, hence, the Jacobian of $\mathbf{r}_{\mathcal{C}_{ik}}$ w.r.t. $\tilde{\delta} \mathbf{b}_{a_i}$ is zero.
Setting $\mathbf{p}^{c_i}_k = [X \ Y \ Z]^T$, the Jacobians of $\mathbf{r}_{\mathcal{C}_{ik}}$ w.r.t. the system states are:
\begingroup
\renewcommand*{\arraystretch}{1.2}  
\setlength{\arraycolsep}{2.0pt}     
\begin{equation}\label{partial_rCik_system_states}
  \frac{\partial \mathbf{r}_{\mathcal{C}_{ik}}}{\partial (\cdot)} = \frac{\partial \mathbf{r}_{\mathcal{C}_{ik}}}{\partial \mathbf{p}^{c_i}_k} \frac{\partial \mathbf{p}^{c_i}_k}{\partial (\cdot)} = \!- \frac{1}{Z} \! \left[ \begin{array}{ccc}
      f_x & 0 & - f_x \!\cdot\! X / Z  \\[1mm]
      0 & f_y & - f_y \!\cdot\! Y / Z
    \end{array} \right] \!\cdot \frac{\partial \mathbf{p}^{c_i}_k}{\partial(\cdot)},
\end{equation}
\endgroup
where $f_x$ and $f_y$ are the focal length of the camera. ${\partial \mathbf{r}_{\mathcal{C}_{ik}}} / {\partial \mathbf{p}^{c_i}_k}$  is derived from the projection function of a pinhole camera model.

Since $\mathbf{p}^{c_i}_k$ is linear in $\mathbf{v}^w_{b_i}$ and $\mathbf{p}^w_k$, and the retraction is simply a vector sum, the Jacobians of $\mathbf{p}^{c_i}_k$ w.r.t. $\delta \mathbf{v}^w_{b_i}$, $\delta \mathbf{p}^w_k$ are simply the matrix coefficients of $\mathbf{v}^w_{b_i}$ and $\mathbf{p}^w_k$. Therefore, we can focus on the following remaining Jacobians:
\begin{flalign}
&\mathbf{p}^{c_i}_k(\mathbf{p}^w_{b_i}+\mathbf{R}^w_{b_i} \delta \mathbf{p}_{b_i}) \notag\\[2mm]
&= \mathbf{R}^c_b  \mathrm{Exp}(\tilde{\omega}_{b_i} t_d) {\mathbf{R}^w_{b_i}}^T  \left( C - \mathbf{R}^w_{b_i} \delta \mathbf{p}_{b_i}  \right) + \mathbf{p}^c_b \notag\\[2mm]
&= \mathbf{p}^{c_i}_k(\mathbf{p}^w_{b_i}) - \mathbf{R}^c_b \mathrm{Exp}(\tilde{\omega}_{b_i} t_d) \delta \mathbf{p}_{b_i} \label{partial_pcik_pwb},  & 
\end{flalign}
\begin{flalign}
&\mathbf{p}^{c_i}_k( \mathbf{R}^w_{b_i} \mathrm{Exp}(\delta \phi_{b_i}) ) \notag\\[2mm]
&= \mathbf{R}^c_b  \mathrm{Exp}(\tilde{\omega}_{b_i} t_d) \mathrm{Exp}(-\delta \phi_{b_i})  {\mathbf{R}^w_{b_i}}^T  C + \mathbf{p}^c_b \notag\\[2mm]
& \!\! \overset{(\ref{exponential_map_first_order_appro})}{\approx}  \mathbf{R}^c_b \mathrm{Exp}(\tilde{\omega}_{b_i} t_d) (\mathbf{I} - {\delta \phi_{b_i}}^\wedge) {\mathbf{R}^w_{b_i}}^T C + \mathbf{p}^c_b \notag\\[2mm]
&= \mathbf{p}^{c_i}_k(\mathbf{R}^w_{b_i}) + \mathbf{R}^c_b \mathrm{Exp}(\tilde{\omega}_{b_i} t_d) {\left({\mathbf{R}^w_{b_i}}^T C \right)}^\wedge \delta \phi_{b_i}  \label{partial_pcik_Rwb},  & 
\end{flalign}
\begin{flalign}
&\mathbf{p}^{c_i}_k( \mathbf{p}^b_c + \mathbf{R}^b_c \delta \mathbf{p}^b_c) \notag\\[2mm]
&= \mathbf{R}^c_b \mathrm{Exp}(\tilde{\omega}_{b_i} t_d) {\mathbf{R}^w_{b_i}}^T  C - {\mathbf{R}^b_c}^T ( \mathbf{p}^b_c + \mathbf{R}^b_c \delta \mathbf{p}^b_c )  \notag\\[2mm]
&= \mathbf{p}^{c_i}_k( \mathbf{p}^b_c) - \delta \mathbf{p}^b_c \label{partial_pcik_pbc}, & 
\end{flalign}
\begin{flalign}
&\mathbf{p}^{c_i}_k( \mathbf{R}^b_c \mathrm{Exp}(\delta \phi^b_c)  ) \notag\\[2mm]
&=  \mathrm{Exp}(-\delta \phi^b_c)\mathbf{R}^c_b \mathrm{Exp}(\tilde{\omega}_{b_i} t_d) {\mathbf{R}^w_{b_i}}^T   C + \mathrm{Exp}(-\delta \phi^b_c) \mathbf{p}^c_b  \notag\\[2mm]
& \!\! \overset{(\ref{exponential_map_first_order_appro})}{\approx} \left( \mathbf{I} - (\delta \phi^b_c)^\wedge \right) \mathbf{p}^{c_i}_k( \mathbf{R}^b_c  )  \notag\\[2mm]
&= \mathbf{p}^{c_i}_k( \mathbf{R}^b_c) + \left( \mathbf{p}^{c_i}_k( \mathbf{R}^b_c) \right)^\wedge \delta \phi^b_c  \label{partial_pcik_Rbc}, &
\end{flalign}
\begin{flalign}
&\mathbf{p}^{c_i}_k( t_d + \delta t_d  ) \notag\\[2mm]
&=  \mathbf{R}^c_b  \mathrm{Exp}(\tilde{\omega}_{b_i}  t_d + \tilde{\omega}_{b_i}  \delta t_d)  {\mathbf{R}^w_{b_i}}^T \!\! \left( C + \mathbf{v}^w_{b_i}  \delta t_d \right) + \mathbf{p}^c_b  \notag\\[2mm]
& \!\! \overset{(\ref{additive_operation_of_lie_algebra})}{ \underset{(\ref{exponential_map_first_order_appro})}{\approx} } \mathbf{R}^c_b \mathrm{Exp}(\tilde{\omega}_{b_i} t_d) \left(\mathbf{I} + \left(\mathbf{J}_r (\tilde{\omega}_{b_i}  t_d) \tilde{\omega}_{b_i} \delta t_d \right)^\wedge \right)  {\mathbf{R}^w_{b_i}}^T  \notag \\[2mm]
& \ \ \  \cdot\! \left( C + \mathbf{v}^w_{b_i}  \delta t_d \right) + \mathbf{p}^c_b  \notag
\\[2mm]
&\approx   \mathbf{p}^{c_i}_k(t_d) + \mathbf{R}^c_b \mathrm{Exp}(\tilde{\omega}_{b_i} t_d)  \notag \\[2mm]
& \ \ \  \cdot\! \left( {\mathbf{R}^w_{b_i}}^T \mathbf{v}^w_{b_i} + \left(\mathbf{J}_r(\tilde{\omega}_{b_i} t_d) \!\cdot\! \tilde{\omega}_{b_i} \right)^\wedge {\mathbf{R}^w_{b_i}}^T C \right) \delta t_d
\label{partial_pcik_td}, &
\end{flalign}
\begin{flalign}
&\mathbf{p}^{c_i}_k( \delta \mathbf{b}_{g_i} + \tilde{\delta} \mathbf{b}_{g_i}  ) \notag\\[2mm]
&=  \mathbf{R}^c_b  \mathrm{Exp}(\tilde{\omega}_{b_i}  t_d - \tilde{\delta} \mathbf{b}_{g_i} t_d)  {\mathbf{R}^w_{b_i}}^T C + \mathbf{p}^c_b  \notag\\[2mm]
& \!\! \overset{(\ref{additive_operation_of_lie_algebra})}{ \underset{(\ref{exponential_map_first_order_appro})}{\approx} } \mathbf{R}^c_b \mathrm{Exp}(\tilde{\omega}_{b_i} t_d) (\mathbf{I} - (\mathbf{J}_r (\tilde{\omega}_{b_i}  t_d) \tilde{\delta} \mathbf{b}_{g_i}  t_d )^\wedge ) {\mathbf{R}^w_{b_i}}^T C + \mathbf{p}^c_b   \notag \\[2mm]
&=  \mathbf{p}^{c_i}_k(\delta \mathbf{b}_{g_i}) + \mathbf{R}^c_b \mathrm{Exp}(\tilde{\omega}_{b_i} t_d)  \left({\mathbf{R}^w_{b_i}}^T C \right)^\wedge  \mathbf{J}_r(\tilde{\omega}_{b_i} t_d)  \tilde{\delta} \mathbf{b}_{g_i} t_d
\label{partial_pcik_td}, &
\end{flalign}
where we used the shorthand $C\doteq \mathbf{p}^w_k  \!- \mathbf{p}^w_{b_i} + \mathbf{v}^w_{b_i} t_d$, and $\tilde{\omega}_{b_i} = \omega_{b_i} - \bar{\mathbf{b}}_{g_i} - \delta \mathbf{b}_{g_i} $. $\mathbf{J}_r (\tilde{\omega}_{b_i} t_d)$ is the right-jacobian matrix of $\tilde{\omega}_{b_i} t_d$.
Summarizing, the Jacobians of $\mathbf{p}^{c_i}_k$ are:
\begin{equation}\label{Jacoobians_of_p_ci_k}
\begin{aligned}
\frac{\partial \mathbf{p}^{c_i}_k}{\partial \delta \mathbf{p}_{b_i}} &= - \mathbf{R}^c_b \mathrm{Exp}(\tilde{\omega}_{b_i} t_d), \ \ \ \ \ \ \ \ \ \ \ \ \ \  \frac{\partial \mathbf{p}^{c_i}_k}{\partial \tilde{\delta} \mathbf{b}_{a_i}} = \mathbf{0}_{3\times3},  \notag\\[1mm]
\frac{\partial \mathbf{p}^{c_i}_k}{\partial \delta \mathbf{v}_{b_i}} &= \mathbf{R}^c_b \mathrm{Exp}(\tilde{\omega}_{b_i} t_d) {\mathbf{R}^w_{b_i}}^T t_d,
\ \ \ \ \ \ \ \ \ \frac{\partial \mathbf{p}^{c_i}_k}{\partial \delta \phi^b_c} = \left(\mathbf{p}^{c_i}_k \right)^\wedge, \notag\\[1mm]
\frac{\partial \mathbf{p}^{c_i}_k}{\partial \delta \phi_{b_i}} &= \mathbf{R}^c_b \mathrm{Exp}(\tilde{\omega}_{b_i} t_d) {\left({\mathbf{R}^w_{b_i}}^T C \right)}^\wedge,
\ \ \ \   \frac{\partial \mathbf{p}^{c_i}_k}{\partial \delta \mathbf{p}^b_c} = -\mathbf{I}_{3\times3}, \notag\\[1mm]
\frac{\partial \mathbf{p}^{c_i}_k}{\partial \delta \mathbf{p}^w_{k}} &= \mathbf{R}^c_b   \mathrm{Exp}(\tilde{\omega}_{b_i} t_d) {\mathbf{R}^w_{b_i}}^T,  \notag\\[1mm]
\frac{\partial \mathbf{p}^{c_i}_k}{\partial \tilde{\delta} \mathbf{b}_{g_i}} &=  \mathbf{R}^c_b \mathrm{Exp}(\tilde{\omega}_{b_i} t_d)  \left({\mathbf{R}^w_{b_i}}^T C \right)^\wedge  \mathbf{J}_r(\tilde{\omega}_{b_i} t_d)  t_d,\notag\\[1mm]
\frac{\partial \mathbf{p}^{c_i}_k}{\partial \delta t_d} &= \mathbf{R}^c_b   \mathrm{Exp}(\tilde{\omega}_{b_i} t_d) \! \left( \! {\mathbf{R}^w_{b_i}}^T \! \mathbf{v}^w_{b_i} \!+\! \left(\mathbf{J}_r(\tilde{\omega}_{b_i} t_d) \!\cdot\! \tilde{\omega}_{b_i} \! \right)^\wedge \!{\mathbf{R}^w_{b_i}}^T \! C \right).  \notag\\[1mm]
\end{aligned}
\end{equation}


\subsection{Jacobians of Rotation Errors}
\label{appendix: Jacobians of the first process}
In this section, we give the Jacobians of rotation error $\mathbf{e}_{rot}$ between consecutive keyframes $i$ and $j$ (here we use $j$ to replace $i+1$ for simplification, see (\ref{e_Rbc_td_gyroBias})) w.r.t. the vectors $\delta \phi^b_c$, $\tilde{\delta} \mathbf{b}_{g_i}$, and $\delta t_d$.

Letting $\phi'_1 =  \mathrm{Log} ((\Delta \bar{\mathbf{R}}_{i,j} \mathrm{Exp}(\mathbf{J}_{\Delta \bar{\mathbf{R}}}^g \delta \mathbf{b}_g) )^T )$,
$ \phi'_2 = \mathrm{Log}(\mathbf{R}'_2) = \mathrm{Log} ( \mathrm{Exp}(-\omega_{c_i} t_d) \mathbf{R}^{c_i}_w\mathbf{R}^w_{c_{j}} \mathrm{Exp}(\omega_{c_{j}} t_d))$,
and $\phi'_3 = \mathrm{Log} ( \Delta \bar{\mathbf{R}}_{i,j}^T \mathbf{R}^b_c \mathbf{R}'_2 \mathbf{R}^c_b )$,
 we have:
\begin{flalign}
    &\mathbf{e}_{rot} (\mathbf{R}^b_c \mathrm{Exp}(\delta \phi^b_c)) \notag \\[2mm]
    & \!\! \overset{(\ref{adjoint_property})}{=} \mathrm{Log}( \mathrm{Exp}(\phi'_1) \mathrm{Exp}(\mathbf{R}^b_c \mathrm{Exp}(\delta \phi^b_c) \phi'_2  ) )  \notag \\[2mm]
    & \!\! \overset{(\ref{exponential_map_first_order_appro})}{\approx}  \mathrm{Log}( \mathrm{Exp}(\phi'_1) \mathrm{Exp}(\mathbf{R}^b_c \phi'_2 + \mathbf{R}^b_c {\delta \phi^b_c}^\wedge \phi'_2 ) )  \notag \\[2mm]
    & \!\! \overset{(\ref{additive_operation_of_lie_algebra})}{\approx}\mathrm{Log} \left(  \mathrm{Exp}(\phi'_1) \mathrm{Exp}(\mathbf{R}^b_c \phi'_2) \mathrm{Exp}\left( \mathbf{J}_r(\mathbf{R}^b_c \phi'_2)  \mathbf{R}^b_c {\delta \phi^b_c}^\wedge \phi'_2 \right)  \right) \notag \\[2mm]
    & = \mathrm{Log} \left(  \mathrm{Exp}\left(\mathbf{e}_{rot} (\mathbf{R}^b_c) \right) \mathrm{Exp}\left( - \mathbf{J}_r(\mathbf{R}^b_c \phi'_2)  \mathbf{R}^b_c {\phi'_2}^\wedge \delta \phi^b_c  \right)  \right) \notag \\[2mm]
    & \! \overset{(\ref{BCH_linear_approximation})} {\approx } \mathbf{e}_{rot} (\mathbf{R}^b_c) - \mathbf{J}_r^{-1}(\mathbf{e}_{rot} (\mathbf{R}^b_c))  \mathbf{J}_r(\mathbf{R}^b_c \phi'_2)  \mathbf{R}^b_c {\phi'_2}^\wedge \delta \phi^b_c     \label{partial_e_rot_Rbc},  & 
\end{flalign}
\begin{flalign}
    &\mathbf{e}_{rot} (\delta \mathbf{b}_{g_i} + \tilde{\delta} \mathbf{b}_{g_i}) \notag \\[2mm]
    & = \mathrm{Log} \left( \mathrm{Exp}\left(-\mathbf{J}_{\Delta \bar{\mathbf{R}}}^g \!\cdot\! \left(\delta \mathbf{b}_{g_i} + \tilde{\delta} \mathbf{b}_{g_i}\right) \right) \mathrm{Exp}\left(\phi'_3\right)  \right)  \notag \\[2mm]
    & \!\! \overset{(\ref{additive_operation_of_lie_algebra})}{\approx} \mathrm{Log}\left( \mathrm{Exp}\left( -\mathbf{J}_l^{b_g} \mathbf{J}_{\Delta \bar{\mathbf{R}}}^g \tilde{\delta} \mathbf{b}_{g_i} \right) \mathrm{Exp} \left( - \mathbf{J}_{\Delta \bar{\mathbf{R}}}^g \delta \mathbf{b}_{g_i} \right) \mathrm{Exp}\left(\phi'_3 \right) \right) \notag\\[2mm]
    & = \mathrm{Log}\left( \mathrm{Exp}\left( -\mathbf{J}_l^{b_g} \mathbf{J}_{\Delta \bar{\mathbf{R}}}^g \tilde{\delta} \mathbf{b}_{g_i} \right) \mathrm{Exp} \left( \mathbf{e}_{rot} (\delta \mathbf{b}_{g_i}) \right)  \right)
    \notag\\[2mm]
    & \! \overset{(\ref{BCH_linear_approximation})}{\approx} \mathbf{e}_{rot} (\delta \mathbf{b}_{g_i}) - \mathbf{J}_l^{-1} \left( \mathbf{e}_{rot} (\delta \mathbf{b}_{g_i})\right) \mathbf{J}_l^{b_g} \mathbf{J}_{\Delta \bar{\mathbf{R}}}^g \tilde{\delta} \mathbf{b}_{g_i}
        \label{partial_e_rot_bg},  & 
\end{flalign}
where we use the shorthand $\mathbf{J}_l^{b_g} \doteq \mathbf{J}_l \left(- \mathbf{J}_{\Delta \bar{\mathbf{R}}}^g \delta \mathbf{b}_{g_i} \right)$.

Letting $\mathbf{R}''_1 = \left(\Delta \bar{\mathbf{R}}_{ij} \mathrm{Exp}\left(\mathbf{J}_{\Delta \bar{\mathbf{R}}}^g \delta \mathbf{b}_g\right) \right)^T \!  \mathbf{R}^b_c$, $\mathbf{R}''_2 = \mathbf{R}^{c_i}_w\mathbf{R}^w_{c_{j}} $, and $\mathbf{R}''_3 = \mathbf{R}^c_b$, we have:
\begin{flalign}
    &\mathbf{e}_{rot} (t_d + \delta t_d) \notag \\[2mm]
    & = \mathrm{Log} \!\left( \mathbf{R}''_1 \mathrm{Exp}(-\omega_{c_i} (t_d \!+ \delta t_d) ) \mathbf{R}''_2 \mathrm{Exp}(\omega_{c_j} (t_d \!+ \delta t_d) ) \mathbf{R}''_3  \right)  \notag \\[2mm]
    & \!\! \overset{(\ref{additive_operation_of_lie_algebra})}{\approx} \mathrm{Log} \left( \mathbf{R}''_1 \mathrm{Exp}(-\mathbf{J}_l^i \omega_{c_i} \delta t_d) \mathrm{Exp}(-\omega_{c_i} t_d) \mathbf{R}''_2 \right. \notag \\[2mm]
      & \left. \ \ \ \  \cdot \mathrm{Exp}(\omega_{c_j} t_d) \mathrm{Exp}(\mathbf{J}_r^j \omega_{c_j} \delta t_d) \mathbf{R}''_3 \right) \notag \\[2mm]
    & \!\! \overset{(\ref{adjoint_property})}{=} \mathrm{Log} \left(  \mathrm{Exp}(- \mathbf{R}''_1 \mathbf{J}_l^i \omega_{c_i} \delta t_d) \mathbf{R}''_1 \mathrm{Exp}(-\omega_{c_i} t_d) \mathbf{R}''_2 \right. \notag \\[2mm]
      & \left. \ \ \ \  \cdot \mathrm{Exp}(\omega_{c_j} t_d) \mathbf{R}''_3 \mathrm{Exp}( {\mathbf{R}''_3}^T \mathbf{J}_r^j \omega_{c_j} \delta t_d)  \right) \notag \\[2mm]
    & = \mathrm{Log} \left(  \mathrm{Exp}(- \mathbf{R}''_1 \mathbf{J}_l^i \omega_{c_i} \delta t_d)
    \mathrm{Exp}(\mathbf{e}_{rot}(t_d)) \right. \notag \\[2mm]
      & \left. \ \ \ \cdot \mathrm{Exp}( {\mathbf{R}''_3}^T \mathbf{J}_r^j \omega_{c_j} \delta t_d) \right) \notag \\[2mm]
    & \!\! \overset{(\ref{adjoint_property})}{=} \mathrm{Log} \big( \mathrm{Exp}(\mathbf{e}_{rot}(t_d)) \mathrm{Exp}(- {\mathrm{Exp}(\mathbf{e}_{rot}(t_d))}^T \mathbf{R}''_1 \mathbf{J}_l^i \omega_{c_i} \delta t_d)  \notag \\[2mm]
      &  \ \ \  \cdot \mathrm{Exp}( {\mathbf{R}''_3}^T \mathbf{J}_r^j \omega_{c_j} \delta t_d)  \big) \notag \\[2mm]
    & = \mathrm{Log} \left( \mathrm{Exp}(\mathbf{e}_{rot}(t_d)) \mathrm{Exp}(D \!\cdot\! \delta t_d) \mathrm{Exp}(E \!\cdot\! \delta t_d) \right) \notag \\[2mm]
    & \!\! \overset{(\ref{exponential_map_first_order_appro})}{\approx} \mathrm{Log} \left( \mathrm{Exp}(\mathbf{e}_{rot}(t_d)) (\mathbf{I} + (D+E)^\wedge \delta t_d) \right) \notag \\[2mm]
    & \!\! \overset{(\ref{exponential_map_first_order_appro})}{\approx} \mathrm{Log} \left( \mathrm{Exp}(\mathbf{e}_{rot}(t_d)) \mathrm{Exp}((D+E) \delta t_d) \right) \notag \\[2mm]
    &\!\! \overset{(\ref{BCH_linear_approximation})}{\approx} \mathbf{e}_{rot}(t_d) + \mathbf{J}_r^{-1}(\mathbf{e}_{rot}(t_d)) ((D+E) \delta t_d)
        \label{partial_e_rot_td},  & 
\end{flalign}
with $\mathbf{J}_l^i \doteq \mathbf{J}_l (-\omega_{c_i} t_d)$, $\mathbf{J}_r^j \doteq \mathbf{J}_r (\omega_{c_j} t_d)$, $D \doteq - {\mathrm{Exp}(\mathbf{e}_{rot}(t_d))}^T \mathbf{R}''_1 \mathbf{J}_l^i \omega_{c_i} $, and $E \doteq  {\mathbf{R}''_3}^T \mathbf{J}_r^j \omega_{c_j} $.
Summarizing, the Jacobians of $\mathbf{e}_{rot}$ are:
\begin{equation}\label{Jacobians_of_e_rot}
\begin{aligned}
\frac{\partial \mathbf{e}_{rot}}{\partial \delta \phi^b_c} &= - \mathbf{J}_r^{-1}(\mathbf{e}_{rot} (\mathbf{R}^b_c))  \mathbf{J}_r(\mathbf{R}^b_c \phi'_2)  \mathbf{R}^b_c {\phi'_2}^\wedge  &\ \notag \\[1mm]
\frac{\partial \mathbf{e}_{rot}}{\partial \tilde{\delta} \mathbf{b}_{g_i}} &= - \mathbf{J}_l^{-1} \left( \mathbf{e}_{rot} (\delta \mathbf{b}_{g_i})\right) \mathbf{J}_l^b \mathbf{J}_{\Delta \bar{\mathbf{R}}}^g ,  &\ \notag \\[1mm]
\frac{\partial \mathbf{e}_{rot}}{\partial \delta t_d} &= \mathbf{J}_r^{-1}(\mathbf{e}_{rot}(t_d)) (D+E) . &\ \notag
\end{aligned}
\end{equation}


%
%

%
%

\ifCLASSOPTIONcaptionsoff
  \newpage
\fi



%

\clearpage

\bibliographystyle{IEEEbib}
\bibliography{refs}

\end{document}